# Semi-supervised reference-based sketch extraction using a contrastive learning framework

Chang Wook Seo
Visual Media Lab, KAIST
South Korea
lgtwins@kaist.ac.kr

Amirsaman Ashtari
Visual Media Lab, KAIST
South Korea
a.s.ashtari@kaist.ac.kr

Junyong Noh
Visual Media Lab, KAIST
South Korea
lgtwins@kaist.ac.kr

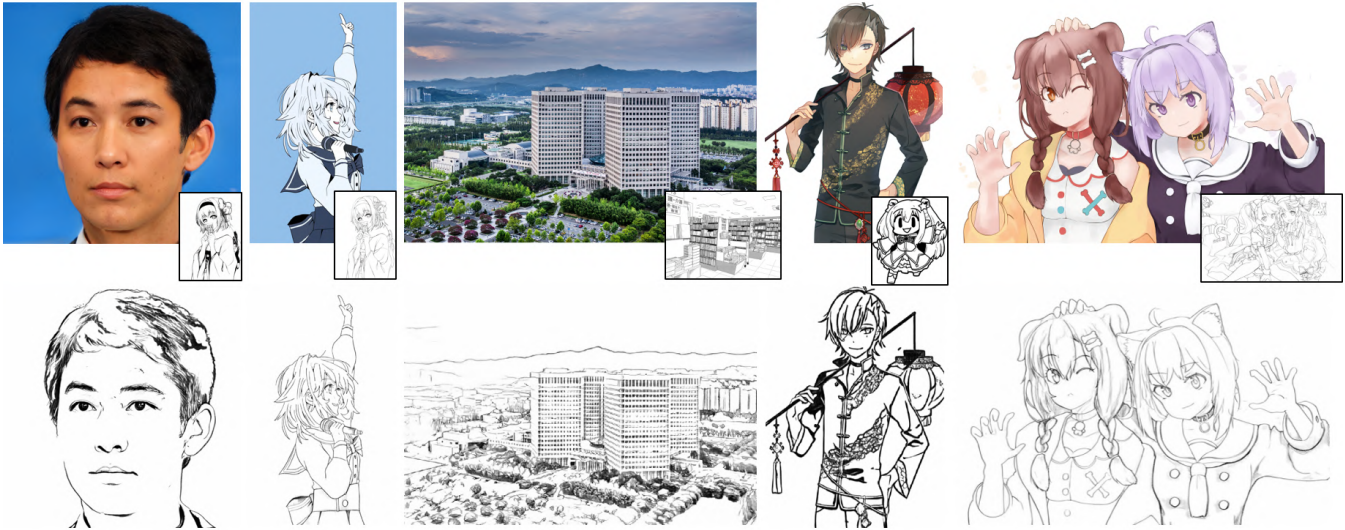

Figure 1: Input color images and sketches extracted by our method. Without requiring repetitive training of the network to make pre-trained weights for each style, our model produces various style sketches by imitating the input reference sketches.© 4SKST (1,2,4), DICC (3), Comet_atr (5)

## ABSTRACT

Sketches reflect the drawing style of individual artists; therefore, it is important to consider their unique styles when extracting sketches from color images for various applications. Unfortunately, most existing sketch extraction methods are designed to extract sketches of a single style. Although there have been some attempts to generate various style sketches, the methods generally suffer from two limitations: low quality results and difficulty in training the model due to the requirement of a paired dataset. In this paper, we propose a novel multi-modal sketch extraction method that can imitate the style of a given reference sketch with unpaired data training in a semi-supervised manner. Our method outperforms state-of-the-art sketch extraction methods and unpaired image translation methods in both quantitative and qualitative evaluations.

## 1 INTRODUCTION

Sketches can be used for a variety of different purposes. Sometimes a sketch can be art by itself and other times it provides a glimpse of the final drawing as an intermediate step. They can also be used to deliver the thoughts of artists as an effective medium for visual communication. For example, artists draw a sketch with relatively complex and thick lines to convey strong impressions. If harmony and balance are intended between the sketch and colors, use of thin and abstract lines is often preferred.

Many computer vision and graphics studies have attempted to automatically extract sketches from photos [Ashtari et al. 2022; Chan et al. 2022] or generating in abstracted lines [Mo et al. 2021; Vinker et al. 2022]. During the extraction process, it is important to infuse the extracted sketches with the style of the authentic drawings, which would have been produced by artists for the outcome to be aesthetically pleasing. From this perspective, most widely used sketch extraction or edge-detection techniques such as Canny [1986] and XDoG [Winnemöller 2011] do not serve the purpose as their results are often noisy or consist of dotted lines.

To address this, SketchKeras [lllyasviel 2017] utilizes deep learning to generate pencil stroke style sketches that closely imitate artistic sketches. Similarly, Chan et al. [2022] proposed a method that can produce high quality artistic sketch drawings by incorporating the geometry and semantics of a color image. Unfortunately, these approaches focus on generating a single style sketch. Ref2sketch [Ashtari et al. 2022] is a multi-modal method that can extract artistic sketches in various styles. Provided with a reference sketch as an additional input, the extracted Ref2sketch sketch closely reflects the style represented in the reference. However, this approach requires a large number of paired sketch and color image data because it is designed to learn the sketch style in a supervised manner.

In this paper, we propose a new sketch extraction method that learns to imitate the style of a reference sketch in the same way as Ref2sketch. However, by leveraging a pre-trained contrastive model, which was trained on paired data, our method can be trained using unpaired sketch and color image data. This approach enhances the efficiency of the training process in a semi-supervised manner. In addition, incorporating attention concatenation that emphasizes the spatial and channel information of inputs improves the quality of the produced sketch. To reflect the style of a reference sketch effectively, we propose a new sketch style loss that utilizes pre-trained weights. These weights are trained based on contrastive learning with a sketch dataset of various styles. We also adopt a line loss by utilizing the HED [Xie and Tu 2015] method to help the model generate a clear and accurate line shape.

Our contributions can be summarized as follows.

- We propose a novel multi-modal sketch extraction method that can imitate the drawing style of the input reference sketch. Our model is trainable with unpaired sketch and color images in a semi-supervised manner.
- We show how generated sketches can be utilized for related studies such as auto-colorization and sketch style transfer.
- In addition to the code, we provide a new authentic sketch dataset prepared by a professional artist. This dataset can assist in precisely evaluating various sketch extraction models. The dataset consists of one of four different styles of sketch drawings paired to 25 color images. The dataset includes a total number of 100 image pairs.

## 2 RELATED WORK
### 2.1 Sketch extraction

There are many sketch extraction techniques designed to generate corresponding sketch images from color images. Some approaches employ an edge-detection method such as Canny [1986], XDoG [Winnemöller 2011], or HED [Xie and Tu 2015]. Other approaches such as SketchKeras [lllyasviel 2017], Anime2sketch [Xiaoyu Xiang 2021], manga line extraction [Li et al. 2017a] and Sketch-simplifications [Simo-Serra et al. 2018, 2016; Xu et al. 2021] have the specific purpose of achieving high quality sketch images using deep learning. Recently, Chan et al. [2022] proposed a novel sketch extraction network that utilizes the depth and semantic meanings of the color image to visualize high quality sketch line drawings. Ref2sketch [Ashtari et al. 2022] is a multi-modal sketch extraction network that learns to imitate an input reference sketch to generate high quality artistic sketch outputs.

These learning-based methods utilize additional loss functions and layers on top of the network models introduced in general domain image-to-image translation studies to improve the performance specifically in the sketch domain. Our method utilizes an attention concatenation layer as well as a set of new loss functions to produce higher quality sketch images compared to previous studies. In addition, our model is trained with unpaired data in a semi-supervised manner to produce a sketch of the style given in the reference image.

### 2.2 Image-to-image translation

Image-to-image translation methods can be divided into several categories: supervised [Ashtari et al. 2022; Isola et al. 2017; Rott Shaham et al. 2021; Wang et al. 2018b,a], unsupervised [Kim et al. 2017; Nizan and Tal 2020; Park et al. 2020a; Xie et al. 2021; Zhu et al. 2017], single-modal [Isola et al. 2017; Rott Shaham et al. 2021; Wang et al. 2018b,a; Xie et al. 2021], and multi-modal [Choi et al. 2020; Lee et al. 2020b; Nizan and Tal 2020; Park et al. 2020b; Ruan et al. 2019]. Supervised methods require paired data for training the model, while unsupervised methods can be trained with unpaired data. Paired data is valuable but rare, especially for authentic sketches drawn by artists; therefore, unsupervised methods make provide convenience for dataset gathering more convenient.

Single-modal methods generate only one output for a given input, while multi-modal methods produce various outputs from either a single input or multiple additional inputs such as a segmentation map [Ntavelis et al. 2020; Sushko et al. 2020; Tang et al. 2022, 2019], text [Kim and Ye 2021; Li et al. 2020a,b,c; Liu et al. 2020a] or a reference image [He et al. 2018; Huang et al. 2018; Li et al. 2022; Ma et al. 2018a; Park et al. 2020b]. Multi-modal methods can easily be applied to diverse applications that require different style images in the same domain, including interior [Lee et al. 2017; Xu et al. 2017; Zheng et al. 2020], human-pose [Schneider et al. 2022; Si et al. 2018; Siarohin et al. 2018], and face emotions [Luna-Jiménez et al. 2021; Savchenko 2022; Seo et al. 2022]. Because different artists draw sketches in different styles, it is important to consider these style differences in sketch based image manipulation studies [Liu et al. 2021; Qi et al. 2021; Simo-Serra et al. 2016; Xu et al. 2021]. Therefore, multi-modal methods have been adopted in various sketch domain studies including photo-sketch synthesis [Chen and Hays 2018; Gao et al. 2020; Li et al. 2019, 2017b; Liu et al. 2020b; Yi et al. 2019, 2020] and sketch auto-colorization [Ci et al. 2018; Kim et al. 2019; Lee et al. 2020a; Liu et al. 2022; Ma et al. 2018b; Thasarathan and Ebrahimi 2019; Yuan and Simo-Serra 2021; Zhang et al. 2018b,b; Zou et al. 2019].

## 3 METHOD

Our goal is to design a model that extracts a sketch from a given color image while imitating the style of a reference sketch. Because pairs of sketch and color image data are scarce, we choose to train the model with an unpaired dataset. Many previous approaches have relied solely on cycle consistency losses [Zhu et al. 2017] in the generator to preserve visual similarity between unpaired data. In addition to using the cycle consistency loss, we introduce two novel losses. The first is line loss that ensures the shape of the output sketch is similar. to that of the input color image. The second is a sketch style loss that enforces the style of the output sketch to follow that of the reference image. The discriminator $D$ of our network examines if the output of the sketch domain generator $O=G_s(C_i, R_i)$ is in the same domain as that of the reference input $R_i$. Here, $C_i$ represents the input color image. $D$ ensures that $O$ lies in a sketch domain. See Figure 2 for an overview of our network design and the supplementary material for the detailed information of our network.

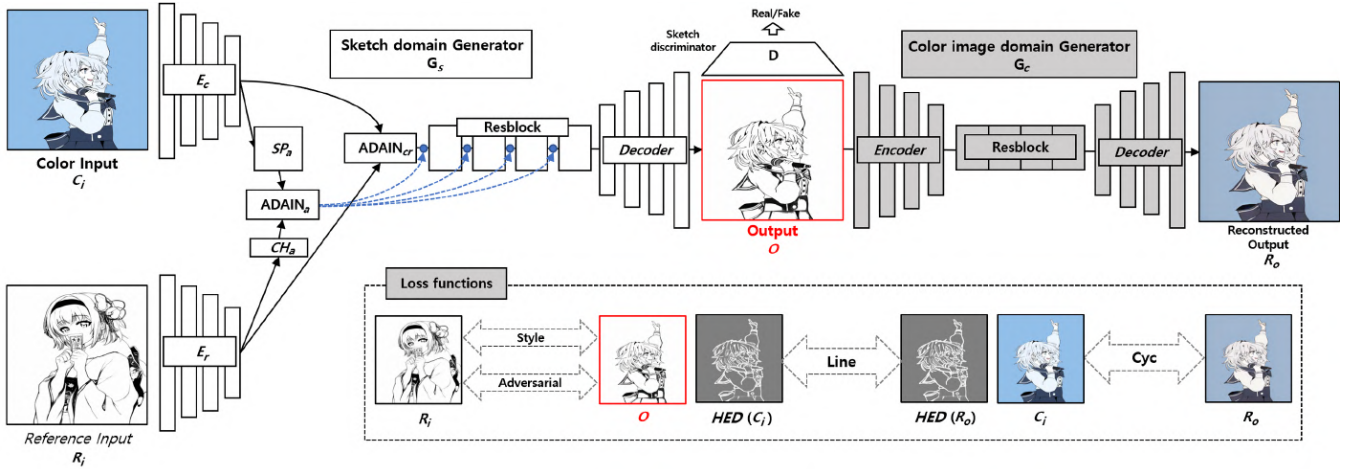

Figure 2: Overview of our network design. See Section 3.1 for the explanation of this network and the definitions for the notations used here. © 4SKST

### 3.1 Overview

Our method performs the following steps to train our sketch-extraction model that produces the output with the style defined by the reference sketch:

(A) Two encoders, $E_c$ and $E_r$ which consist of convolution layers, extract the features from the input color image $C_i$ and input reference image $R_i$, respectively.
(B) The extracted features from $E_c$ are fed into the spatial attention layer $SP_a$ and features from $E_r$ are fed into the channel attention layer $CH_a$. The outputs from the attention layers then go through adaptive instance normalization $ADAIN_a$ [Huang and Belongie 2017].
(C) Simultaneously with step (B), the features from $E_c$ and $E_r$ directly go through another adaptive instance normalization $ADAIN_{cr}$.
(D) Resblock, which consists of 4 convolution block layers, receives the outputs from both $ADAIN_a$ and $ADAIN_{cr}$. The output from $ADAIN_a$ is concatenated to the output from $ADAIN_{cr}$ before the combination is concatenated to the output of each Resblock layer.
(E) The output from Resblock goes through the decoder layers to produce an output sketch.
(F) To preserve the shape of the color image input in the output sketch, the output sketch is fed into the color image domain generator $G_c$. $G_c$, which consists of the encoder-decoder and Resblock layers, produces a reconstructed output $R_o$. We calculate four different loss functions using $O$ and $R_o$ to ensure that the output sketch imitates the style of the reference input while preserving the shape of the color image input.

### 3.2 Attention

Our network includes spatial attention and channel attention to emphasize the shape and style of each input. The attention structures are similar to those of the CBAM [Woo et al. 2018] method. To reflect the shape given by $C_i$, the spatial attention is placed after $E_c$. Likewise, to adopt the style given by $R_i$, the channel attention is placed after $E_r$. The details of the attentions are as follows.

For input feature maps, $E_c(C_i) \in \mathbb{R}^{C \times H \times W}$ and $E_r(R_i) \in \mathbb{R}^{C \times H \times W}$, where $C, H,$ and $W$ represent the number of channels, the height and the width of the image, respectively. We compute the spatial and channel attentions individually (i.e., $SP_a \in \mathbb{R}^{1 \times H \times W}$ and $CH_a \in \mathbb{R}^{C \times 1 \times 1}$) as follows:

$$\begin{aligned}
&SP_a(E_c(C_i)) \\
&= \sigma(f^{3\times 3}([AvgPool^{sp}(E_c(C_i)); MaxPool^{sp}(E_c(C_i))])) \\
&= \sigma(f^{3\times 3}([E_c(C_i)^{sp}_{avg}; E_c(C_i)^{sp}_{max}])),
\end{aligned} \quad (1)$$

$$\begin{aligned}
&CH_a(E_r(R_i)) \\
&= \sigma(MLP(AvgPool^{ch}(E_r(R_i))) + MLP(MaxPool^{ch}(E_r(R_i)))) \\
&= \sigma(W_1(W_0(E_r(R_i)^{ch}_{avg})) + W_1(W_0(E_r(R_i)^{ch}_{max}))),
\end{aligned} \quad (2)$$

In Eq. (1), the features from $E_c$ are pooled by two different pooling functions before convolved with a $3 \times 3$ kernel filter. In Eq. (2), the features from $E_r$ are pooled by two different pooling functions before going through MLP layers. Here, $W_0 \in \mathbb{R}^{C/r \times 1}$ and $W_1 \in \mathbb{R}^{C \times C/r}$ represent the weights of MLP layers, with a reduction ratio $r = 16$. Sigmoid functions $\sigma$ are used for both attentions. The feature sizes after the pooling layers are $AvgPool^{ch}, MaxPool^{ch} \in \mathbb{R}^{C \times 1 \times 1}$ and $AvgPool^{sp}, MaxPool^{sp} \in \mathbb{R}^{1 \times H \times W}$.

These attention features are then multiplied element-wise using the Hadamard product ⊙ with their original input features $E_c(C_i)$ and $E_r(R_i)$ before being normalized by $ADAIN_a$. ADAIN aligns the mean and variance of the features from the color and reference inputs:

$$ADAIN_a(E_c(C_i) \odot SP_a(E_c(C_i)), E_r(R_i) \odot CH_a(E_r(R_i))). \quad (3)$$

These normalized features are concatenated to the features from $ADAIN_{cr}$ at Resblock to make the model trainable at a higher resolution. Training the model at a low resolution such as 256×256 without the attention feature concatenation preserves the shape but produces low-quality sketches. Training the model at a higher resolution such as 512×512 without the concatenation causes shape distortion. See Figure 3 for examples. More discussion on the benefit of this attention design can be found in the supplementary material.

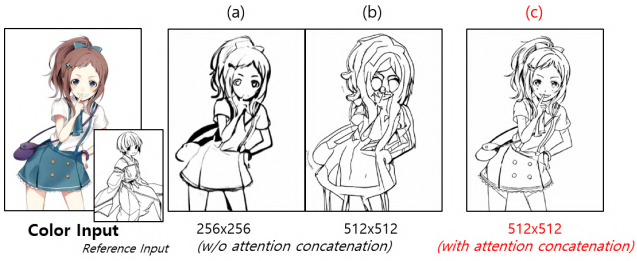

Figure 3: Examples of the output produced with and without attention concatenation in the network architecture. For this experiment, we removed $SP_a$, $CH_a$, and $ADAIN_a$ as well as their connections to Resblock layers. The outputs produced without the attention concatenation, (a) and (b), show poor quality results while the output using the concatenation shows a better result produced at a higher resolution of 512×512. © 4SKST

## 3.3 Losses

*3.3.1 Sketch Style Loss.* Here, we introduce a novel loss function that calculates the style difference in the sketch domain. This loss is computed using weights pre-trained based on contrastive learning [Chen et al. 2020] that employs a triplet loss. Contrastive learning maps similar features closer together and dissimilar features further away in embedded space.

In our method, the model is trained to map sketches of similar styles closer together (Anchor and Positive) and sketches of the same shape but different styles (Negative) further away (as illustrated in Figure 4). The dataset for training the network is obtained using Ref2sketch [Ashtari et al. 2022], which allows sketches of different styles to be generated from a single input image.

To compute the style loss, the reference input $R_i$ and output $O = G_s(C_i, R_i)$ generated by the sketch domain generator $G_s$ are used. Although these two images are of different shapes, they should be of the same style; therefore, we extract the style feature embeddings from them using pre-trained contrastive weights. We then apply L1 normalization to calculate the difference of the two embeddings. The loss is expressed as follows:

$$\mathcal{L}_{style} = \mathbb{E}_{O,R_i}[||C_w(O) - C_w(R_i)||_1]. \quad (4)$$

Symbol $C_w$ represents the pre-trained contrastive learning model that extracts style feature embeddings. Without this style loss function, the network fails to imitate the drawing style of the reference input and consequently produces a fixed style output regardless of the reference input. Figure 5 shows failed examples. Similar to Simo-Serra et al. [2018], this loss function is pre-trained with paired data to enable our main model to be trained with unpaired data. This makes our approach semi-supervised. Refer to the supplementary material for more details regarding the contrastive learning weights used in our method.

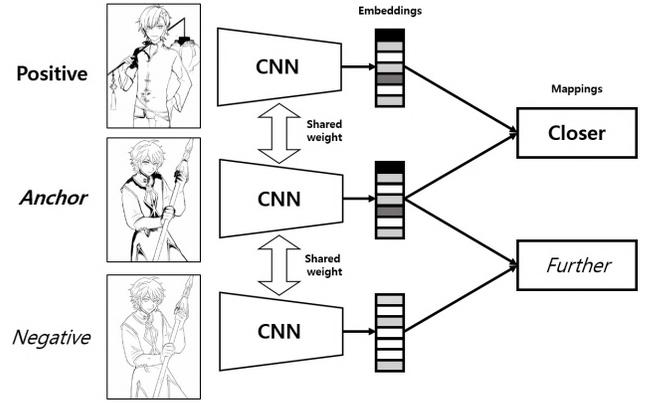

Figure 4: Illustration of contrastive learning, which groups similar sketch styles in embedded space regardless of the image shape. © 4SKST

*3.3.2 Line Loss.* To enforce the shape of the output to be identical to the color image input, we apply a loss function that compares the edges of the reconstructed output $R_o = G_c(O)$ generated by the color domain generator $G_c$ and the color input $C_i$ using the HED [Xie and Tu 2015] method. HED [Xie and Tu 2015] detects the edges from the input image. The loss is expressed as follows:

$$\mathcal{L}_{Line} = \mathbb{E}_{C_i,R_o}[\sum_l ||\phi_l(HED(C_i)) - \phi_l(HED(R_o))||_1], \quad (5)$$

Applying HED to the color input $C_i$ and reconstructed output $R_o$ generates edge-detected images, as shown in Figure 2. The differences between the two edge-detected images, $HED(C_i)$ and $HED(R_o)$, are calculated by the perceptual loss function [Johnson et al. 2016] that is designed to compare images based on the pre-trained VGG16 [Simonyan and Zisserman 2014] model. $\phi_l$ denotes the activation map from the $l^{th}$ layer of the VGG16 network. Without this line loss function, the network produces a shape that looks different from that of the color image input. Figure 5 shows failed examples, particularly on the dog's muzzle. This loss function is inspired by Yi et al. [2020].

*3.3.3 Cycle Consistency Loss.* We enforce the shape similarity further by comparing the overall visual difference between $C_i$ and $R_o$. L1 normalization is used, and the loss is expressed as follows:

$$\mathcal{L}_{Cyc} = \mathbb{E}_{C_i,R_o}[||C_i - R_o||_1], \quad (6)$$

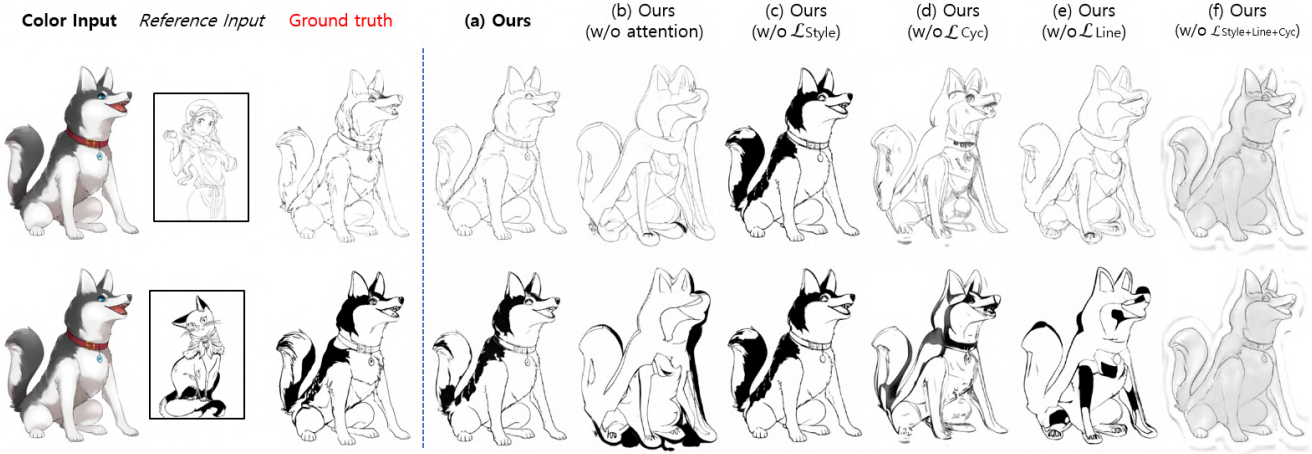

Figure 5: Examples of outputs produced with and without using loss functions. (c) and (f) show the results from the model trained without using the style loss. It is evident that the results have a fixed style regardless of the given reference input. (a), (b), (d), and (e) show the results produced using the style loss. Clearly, the results displayed in the top and bottom rows have different styles according to the reference input. (e) shows the results from the model trained without using the line loss. The shape of the color image input is incorrectly represented. (d) shows the results trained with the line loss but without the cycle consistency loss. The image shapes of (d) are better than (e) but still do not correctly represent the original shape. © 4SKST

### 3.3.4 Adversarial Loss.
The adversarial loss forces the discriminator $D$ to classify the synthetic output $O$ to be in the domain similar to that defined by $R_i$, which is a sketch in our case. The loss is expressed as follows:

$$\mathcal{L}_{adv} = \mathbb{E}_{R_i}[\log(D(R_i))] + \mathbb{E}_O[\log(1 - D(O))], \quad (7)$$

The total loss function for the generator G and discriminator D is defined as follows:

$$\min_G \max_D \mathcal{L}_{total} = \lambda_{style}\mathcal{L}_{style} + \lambda_{line}\mathcal{L}_{line} \\ + \lambda_{cyc}\mathcal{L}_{cyc} + \lambda_{adv}\mathcal{L}_{adv}. \quad (8)$$

The parameters used in Eq. 8 are $\lambda_{style,line} = 5 - \frac{4.5i}{n}$, $\lambda_{cyc} = 10$, and $\lambda_{adv} = 1$, where i is the current epoch number and n is the total number of epochs.

## 4 EXPERIMENT
### 4.1 Experiment setup
*4.1.1 Dataset.* To train our network and baselines, we utilized images from safebooru [DanbooruCommunity 2021]. For sketch domain images, we used images with the tag *line art* in safebooru.

A total of 4,302 sketch domain images are used for training. For the color domain, 3,804 color images were randomly selected from safebooru, which are not used as sketch data. Refer to the supplementary material for detailed information regarding the training dataset.

To evaluate our method and baselines, we created a new sketch dataset of four different styles with the help of a professional artist. Recently, Ref2sketch [Ashtari et al. 2022] evaluated their model and baselines with sketches of four different styles, which consists of a total of 60 pairs. Similarly, we performed the evaluation with a dataset of four different styles to verify how well the output of each model imitates the given reference style. Our **4 sketch style (4SKST) dataset** consists of each of four different style sketches for 25 color images, constituting a total of 100 pairs. Color images were chosen from two different domains: anime-art and real photo. These images and sketches are free to re-distribute for non-commercial purposes (CC-NC)[1]. Four different sketch drawing styles in the dataset follow four major sketch drawing styles. The four major styles were determined by applying K-means clustering [Lloyd 1982] to the same 4,302 sketch images used for training.

*4.1.2 Training details.* An Adam optimizer [Kingma and Ba 2014] with a batch size of four was used in the training. All networks were trained from scratch with a learning rate of 0.0002, and the total number of training epochs was 100 with a constant learning rate for the first 50 epochs followed by a learning rate linearly decayed to zero over the next 50 epochs.

### 4.2 Ablation study
For an ablation study, our network was trained with the attention layers and losses removed according to Figure 5. The evaluation of each model was performed using the 4SKST dataset. When extracting a sketch from a color image input, the same sketch style was used; however, unseen shape images from the dataset were used as the input reference sketch. As evaluation metrics, we employed PSNR [Wang et al. 2004], FID [Heusel et al. 2017], and LPIPS [Zhang et al. 2018a] to compare the distribution of features from the output and ground truth sketches. Table 1 shows the results and Figure 5 illustrates the examples.

The model trained without using attention concatenation performs very poorly and produces the worst results. The models trained without using the style loss, cycle consistency loss, or line loss also produce much different results than the ground truth. The results produced without simultaneously using these three losses have a visual quality far from the sketch domain and achieve the worst scores in both PSNR and FID.

---
[1]CreativeCommons for Non-Commercial uses

### 4.3 Comparison with baselines

We chose six different baselines to compare with our method, MUNIT [Huang et al. 2018], Park et al. [2020b], Ref2sketch [Ashtari et al. 2022], CouncilGAN [Nizan and Tal 2020], IrwGAN [Xie et al. 2021], and Chan et al. [2022]. MUNIT and Park et al. [2020b] are unsupervised multi-modal image translation methods that imitate the reference input for the output style. Note that these methods are designed for general image domain translations. Ref2sketch is the most recent multi-modal method with supervised learning specifically designed for the sketch domain. Council-GAN and IrwGAN are unsupervised image translation methods that solve the limitation of cycle consistency learning based methods [Zhu et al. 2017] by leveraging the collaboration between GANs and the importance reweighting technique. Unfortunately, these methods do not accept a reference input to imitate, and thereby cannot produce a sketch of a desired style. Similarly, while Chan et al. [2022] can convey the semantic and depth meaning of the color image input to an output sketch in an unsupervised manner, it is a single-modal method that can produce the sketch of only one style unless the model is retrained with a dataset of different style sketches.

For comparison, we trained the baseline models with the same dataset described in Sec. 4.1. The training parameters and details were determined based on their official code and the descriptions in their respective papers. Because Ref2sketch [Ashtari et al. 2022] is a supervised method that requires a paired dataset, we utilized pre-trained weights from the official page [ref2sketch 2022]. In the evaluation of each model, methods that can accept a reference image [Ashtari et al. 2022; Huang et al. 2018; Park et al. 2020b] use the same style sketch with an unseen shape image from the 4SKST dataset. Other methods that cannot accept a reference image [Chan et al. 2022; Nizan and Tal 2020; Xie et al. 2021] generate the output based on the color image input. Figure 6 shows examples of the generated outputs. The output sketches produced by these methods were compared to the ground truth with the same three different evaluation metrics [Heusel et al. 2017; Wang et al. 2004; Zhang et al. 2018a] used for the ablation study. A total of 100 pairs of images from the 4SKST dataset were used for each evaluation. Refer to the supplementary material for a detailed explanation regarding the experiment. The results reported in Table 1 confirm that our method outperforms all the baselines in the three different evaluation metrics.

### 4.4 Perceptual study

We further evaluated the performance of our method based on human perceptual judgment. A total of 200 people participated in this study, and a survey consisting of 20 comparisons was created for our evaluation. We provided each participant with a target image and seven results, including ours, in each comparison. We then asked each participant to select the result that looks most similar to the target image. No time constraint was imposed on the participants in this process. Figure 7 shows an example survey and Table 2 lists the resulting scores. More examples can be found in the supplementary material. The result of this perceptual study clearly verifies that our method outperforms the baselines in human perception.

Table 1: Quantitative results from the ablation study and from the comparison with baselines.

| Methods | PSNR↑ | FID↓ | LPIPS↓ |
|---|---|---|---|
| **Ours** | **35.58** | **82.18** | **0.1271** |
| Ours w/o attention | 33.80 | 146.87 | 0.3356 |
| Ours w/o $\mathcal{L}_{Style}$ | 34.96 | 139.28 | 0.1738 |
| Ours w/o $\mathcal{L}_{Cyc}$ | 34.68 | 121.71 | 0.2357 |
| Ours w/o $\mathcal{L}_{Line}$ | 34.24 | 125.15 | 0.2660 |
| Ours w/o $\mathcal{L}_{Style} + \mathcal{L}_{Line} + \mathcal{L}_{Cyc}$ | 33.59 | 157.92 | 0.2598 |
| MUNIT | 34.23 | 144.82 | 0.2582 |
| Park et al. [2020b] | 35.04 | 174.12 | 0.2745 |
| Ref2sketch | 35.02 | 115.96 | 0.2192 |
| Council-GAN | 31.76 | 215.81 | 0.4632 |
| IrwGAN | 35.36 | 125.14 | 0.2229 |
| Chan et al. [2022] | 35.05 | 128.96 | 0.2130 |

Table 2: Results from the user perceptual study

| Method | User Score |
|---|---|
| **Ours** | **79.50%** |
| Ref2sketch | 10.66% |
| Chan et al. [2022] | 4.58% |
| IrwGAN | 3.08% |
| MUNIT | 1.16% |
| Park et al. [2020b] | 1.02% |
| Council-GAN | 0% |

### 4.5 Cyclic evaluation

To prove the superiority of our method in preserving the style of the given reference when extracting sketches compared to other methods, we implemented the cyclic evaluation proposed in Ref2sketch [Ashtari et al. 2022]. The main idea of the cyclic evaluation is that, because the extracted sketch should have a style similar to that of the reference input, using the output as the reference image in turn will lead to the same sketch in the original style. Figure 8 illustrates this process. In this evaluation, we chose MUNIT [Huang et al. 2018], Park et al. [2020b], and Ref2sketch as baselines because these methods are designed to accept a reference input to imitate the style. The 4SKST dataset was used for the evaluation. As shown in Table 3, our method outperforms the baselines in LPIPS, FID, and PSNR scores.

### 4.6 Applications

*4.6.1 Improving auto-colorization.* As described in Sec. 2.2, a sketch extraction method that can produce sketches of various styles can also be used to improve the performance of related applications. For example, sketch auto-colorization models typically require many sketches paired with color images [Kim et al. 2019; Lee et al. 2020a; Liu et al. 2022; Ma et al. 2018b; Thasarathan and Ebrahimi 2019; Yuan and Simo-Serra 2021; Zhang et al. 2018b]. These models can benefit from our method in that the use of a multi-style sketch dataset can help avoid an over-fitting problem caused by relying on the sketches of a single style for training.

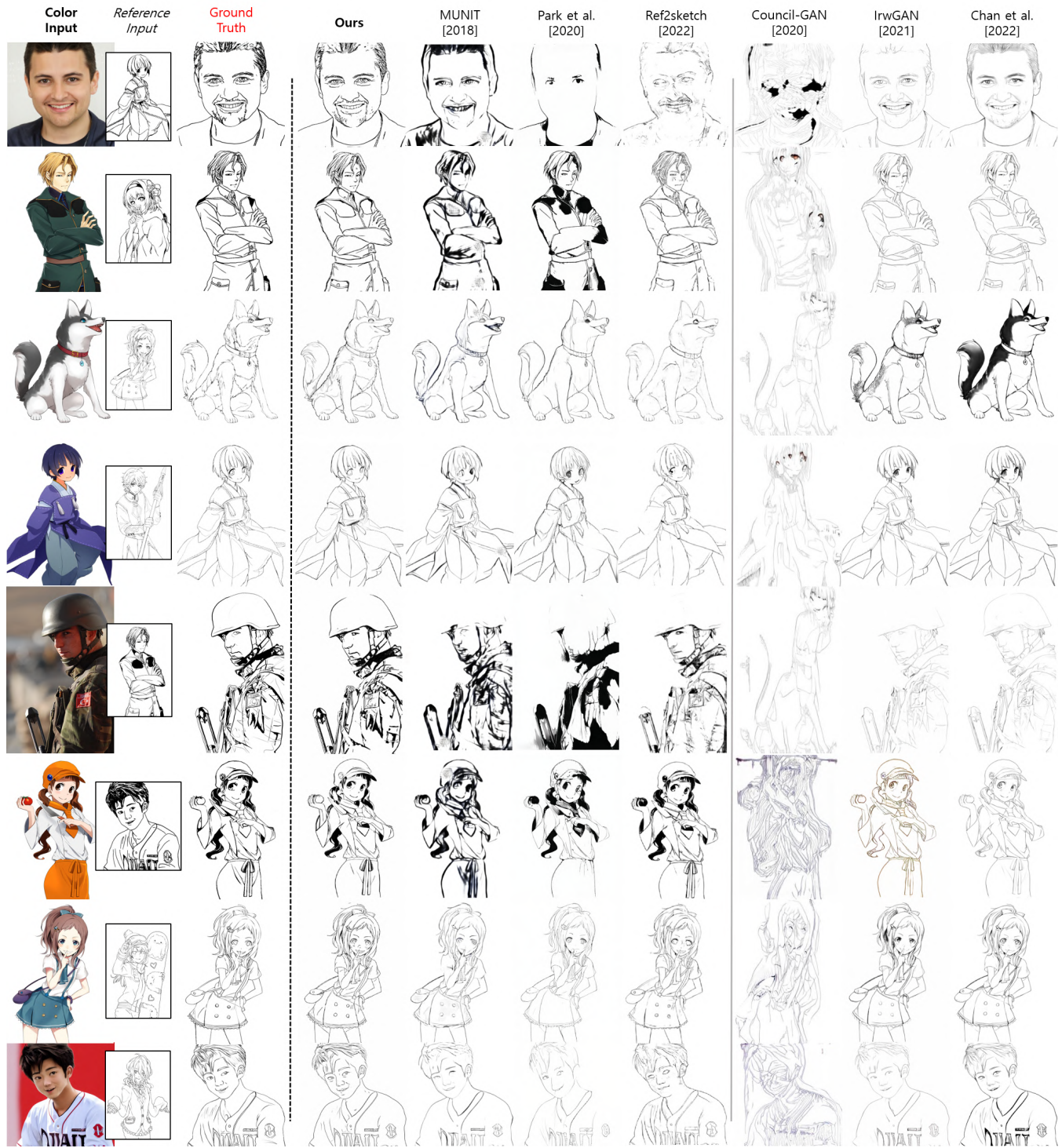

Figure 6: Various examples generated by our method and baselines. While Ref2sketch produces high quality results in some cases as shown in the 6th row, overall our method produces the best quality results in most cases. © 4SKST

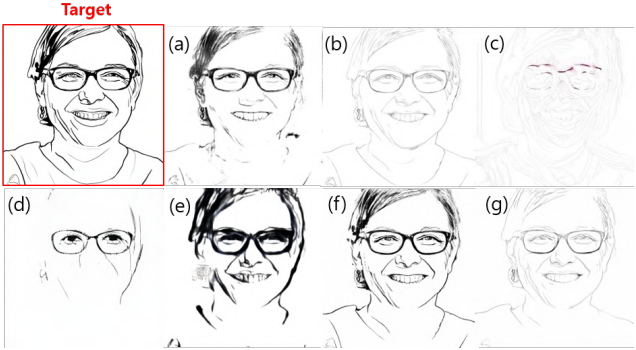

Figure 7: Sample survey prompt for our human perception study. The presented results were produced by our method and six baseline methods: (a) Ref2sketch, (b) Chan et al. [2022], (c) Council-GAN, (d) Park et al. [2020b], (e) MUNIT, (f) Ours, and (g) IrwGAN. The displayed order of these results was chosen randomly for each comparison. © 4SKST

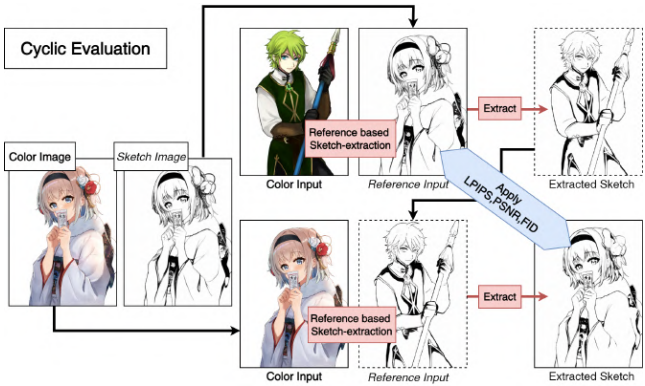

Figure 8: Illustration of the cyclic evaluation proposed in Ref2sketch [Ashtari et al. 2022]. © 4SKST

Table 3: Results of the cyclic evaluation

| Cylclic evaluation | PSNR↑ | FID↓ | LPIPS↓ |
|---|---|---|---|
| **Ours** | **35.84** | **91.97** | **0.1288** |
| Ref2sketch | 35.34 | 130.02 | 0.1643 |
| Park et al. [2020b] | 34.48 | 155.44 | 0.2824 |
| MUNIT | 33.92 | 151.04 | 0.2987 |

To prove this, we trained an auto-colorization network with a paired sketch dataset which was generated by our method and baseline methods. A widely used deep-learning based sketch auto-colorization method [Ci et al. 2018] was chosen, and the model was trained with 1,500 color images from safebooru [DanbooruCommunity 2021], sketches generated by our network, and by the baseline methods used for the experiments performed in auto-colorization papers [Ci et al. 2018; Kim et al. 2019; Thasarathan and Ebrahimi 2019; Yuan and Simo-Serra 2021]. After training the model, we evaluated the quality of the auto-colorized images through authentic sketch inputs and the ground truth color images from the 4SKST dataset. Refer to the supplementary material for more detailed explanations regarding this experiment. Table 4 shows the comparison results.

Table 4: Comparison of the results from the auto-colorization method that was trained using different datasets. The "combined" category consists of sketches extracted by Canny, SketchKeras, XDoG and Simo-Serra et al. [2016].

| Dataset | PSNR↑ | FID↓ | LPIPS↓ |
|---|---|---|---|
| **Ours** | **29.40** | **227.69** | **0.4802** |
| Canny | 28.68 | 268.10 | 0.5892 |
| SketchKeras | 27.99 | 283.69 | 0.6775 |
| XDoG | 28.55 | 239.63 | 0.5931 |
| Simo-Serra et al. [2016] | 28.49 | 270.99 | 0.6295 |
| Combined | 28.78 | 232.37 | 0.5604 |

The colorization model trained with the dataset generated by our method outperformed the others because the sketch dataset with various styles generated by our method helps avoid over-fitting to a specific single sketch style. We acknowledge that a similar effect would be achieved by training the model using the combined data from various sketch extraction methods, as Yuan and Simo-Serra [2021] attempted. These results are represented by "Combined" in Table 4. Our method still produces higher quality colorization results. It is also much simpler to extract multi-style sketches using our method than collecting data from many different methods. See Figure 9 for the colorized image examples.

*4.6.2 Sketch style transfer.* Similar to existing sketch style transfer methods [Liu et al. 2021; Simo-Serra et al. 2018, 2016; Xu et al. 2021], our method can transfer the style of a sketch directly to other sketches without extracting them from color images. This can be very useful when sketch artists work together. In the creation of comics, for example, character sketches are placed on template background sketches. Due to the involvement of many different artists, sketches are often prepared with different styles; therefore, this process requires a manually intensive arrangement of the same style sketches. Automatically transferring the style of a character sketch to match that of the background template will streamline this collaboration process. Figure 10 shows an example of this application. This sketch style transfer application can also be used for the purpose of sketch simplification [Simo-Serra et al. 2018, 2016; Xu et al. 2021]. Digitized rough sketches typically go through a simplification process for the purpose of cleaning up the image. Because it can accept a guiding reference for the simplification, our model can be more instrumental in creating cleaned up sketches in a desired uniform style compared to alternative methods. Refer to Figure 11 and the supplementary material for examples.

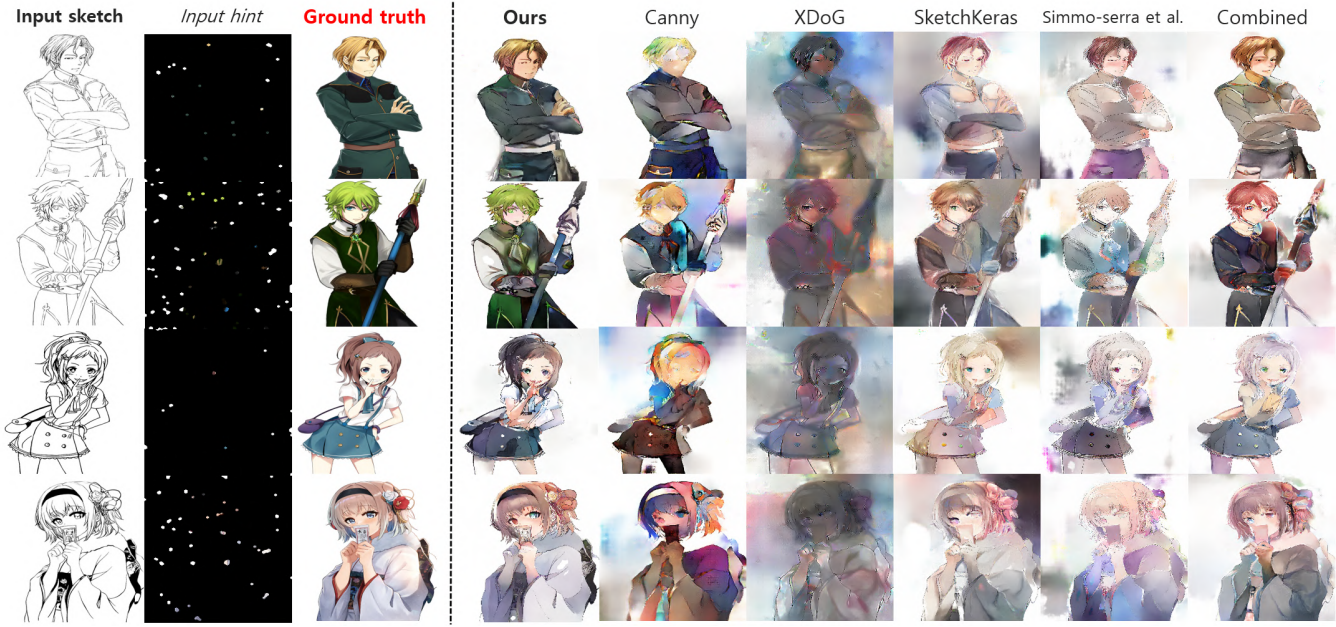

Figure 9: Examples of the output from the auto-colorization method [Ci et al. 2018] trained with different datasets. © 4SKST

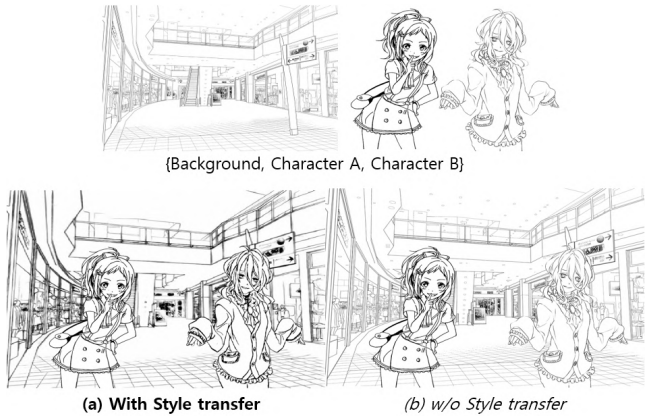

Figure 10: Sketches of different styles can be placed easily in one scene after the style transfer by our method. Visually consistent (a) and inconsistent (b) placements of characters on the template background. © 4SKST (character), tzcat (background)

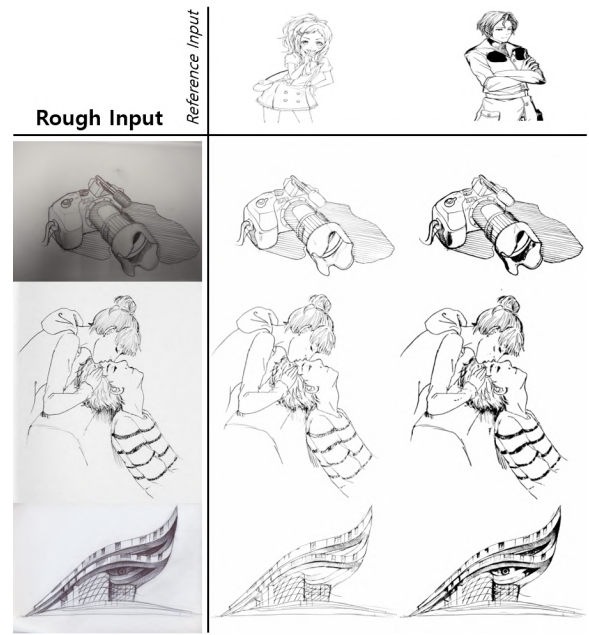

Figure 11: Rough sketches can be simplified by transferring the style with our method. Example rough sketches are from Yan et al. [2020].

## 5 LIMITATIONS AND FUTURE WORK

Our method produces high-quality results when the categories of the color image and the reference image match. For example, higher quality landscape sketches will be extracted from a landscape photo when a reference sketch with landscape content is provided. See Figure 12 for the examples. This unaligned image problem may be alleviated by adopting an importance reweighting method such as that proposed in IrwGAN [Xie et al. 2021]. How to apply this general image domain approach to the sketch domain is not yet clear and may be considered an interesting future research direction.

Our method utilizes sketches extracted by Ref2sketch [Ashtari et al. 2022] when pre-training the contrastive learning model that works as the style loss function of our training network. Therefore,

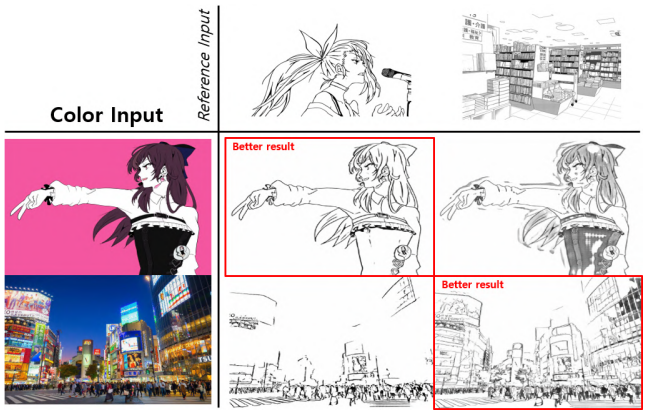

Figure 12: When the categories of the reference and color images match, a higher quality sketch is extracted from the input color image. © 4SKST (character), tzcat (background)

our method shares the same limitation with Ref2sketch. Specifically, our method cannot imitate the style of a reference sketch that does not consist of lines (e.g., pointillism art). The examples in Figure 13 illustrate this. This problem can be addressed by providing a dataset of these styles when pre-training the contrastive learning model.

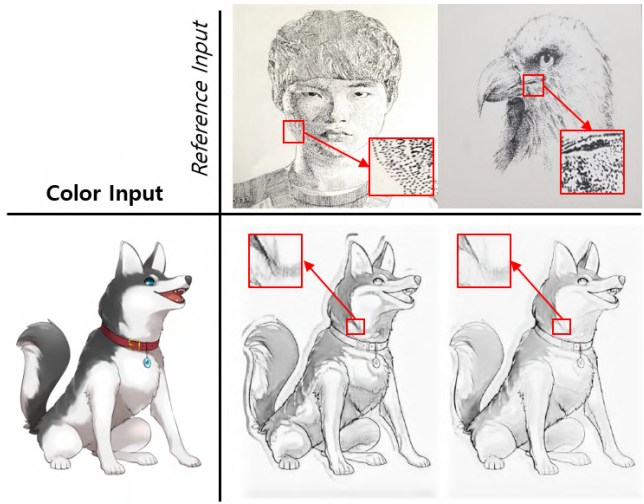

Figure 13: Similar to Ref2sketch, our method fails to extract sketches of the reference style that does not consist of lines, such as pointillism art. © 4SKST (character), Kang (pointillism art)

## 6 CONCLUSION

In this study, we proposed a novel multi-modal method that can extract sketches from a color image in a style given by a reference image. Our method is trained efficiently in a semi-supervised manner. To imitate the style of the reference sketch, we used a pre-trained sketch style loss based on contrastive learning. The pre-training was performed with a paired dataset generated by Ref2sketch [Ashtari et al. 2022], which, in turn, was trained with another paired dataset. Leveraging the previous methods trained with paired data, our method was trained with unpaired color and sketch images.

To preserve the shape of the color input, we introduce a line loss function that is used on top of a cycle consistency loss function. Incorporation of the attention concatenation that emphasizes spatial and channel information enables our model to be trainable in high resolution, producing better quality results than those produced by the baselines. We verified the effectiveness of our method through both quantitative and qualitative experiments. We believe that our method can be utilized in the industry in the form of diverse applications and can stimulate further research related to the generation of various style sketches.

# 1 SUPPLEMENTARY CONTENTS

In the supplementary material, we provide details of our experiments and examples.

The table of contents are as follows:

(1) Section 2, Tables 1, 2 and Figure 1 describe the details of our model including the pre-trained sketch style loss.
(2) Section 3 and Figures 2, 3, 5, 6 describe the details of our 4SKST dataset.
(3) Section 4 describes the details of the comparison.
(4) Section 5, Table 3, and Figure 7 describe the effect of using channel and spatial attention concatenation.
(5) Section 6, Table 3, and Figure 7 describe the details of adding the clip based semantic loss.
(6) Section 7 and Figures 8, 9, 10, 11, 12 describe the details of the perceptual study.
(7) Section 8, Table 4, and Figures 13, 14 describe the details of improving auto-colorization.
(8) Section 9 and Figure 15 describe the details of sketch style transfer.
(9) Section 10 and Figures 16, 17, 18, 19 describe the details of the qualitative comparison to Ref2sketch.
(10) Section 11 and Figures 20, 21 present additional examples of images generated by our method.

# 2 METHOD DETAILS

*Network:* Our network contains two types of encoder, $E_r$ and $E_c$ for encoding the reference style and colorized image, respectively. Both of the encoders have the same architecture presented in Table 1. For the discriminator, we use the PatchGAN architecture with a receptive field of 70 × 70.

*Sketch Style loss :* Our network utilizes a pre-trained model to calculate the style difference between the output sketch and reference input sketch. The pre-trained model is based on contrastive learning as described in the main paper. The details of the model architecture is described in Table 2. To collect a dataset for training the model, we used Ref2sketch [Ashtari et al. 2022] to extract sketches of different styles but the same shape from 1,000 color images. Sketches of one of four different styles are generated using Ref2sketch. While training, Positive image is selected randomly from sketches of the same style as Anchor image, and Negative image is selected randomly from sketches of different styles but the same shape as Anchor image. The training epochs is 200 total and an Adam optimizer [Kingma and Ba 2014] is used with the batch size of 4. The learning rate starts with 0.0002 for first 100 epochs and linearly decays to zero for next 100 epochs.

To locate anchor $f(A)$ and positive embeddings $f(P)$ closer while negative embeddings $f(N)$ farther away, we use a triplet loss to training the model with margin value $a = 1.0$. The loss is expressed as follows:

$$\mathcal{L}_{triplet} = max(||f(A) - f(P)||^2 - ||f(A) - f(N)||^2 + a, 0)] \quad (1)$$

Table 1: In, O, K, P, S, and R denote the number of input channels, the number of output channels, the kernel size, the padding size, the stride size, and the reduction ratio, respectively. Attention functions $SP_c$, $CH_r$ and the feature concatenation before input to resblocks are described in the main paper.

| Sketch domain Generator $G_s$ | |
|---|---|
| **Layer** | $E_c$: **Encoder(color input)** |
| $E_{c1}$ | Conv(In:1, O:64, K:7, P:1, S:1), BNorm, ReLU |
| $E_{c2}$ | Conv(In:64, O:128, K:3, P:1, S:1), BNorm, ReLU |
| $E_{c3}$ | Conv(In:128, O:256, K:3, P:1, S:1), BNorm, ReLU |
| $E_{c4}$ | Conv(In:256, O:512, K:3, P:1, S:1), BNorm, ReLU |
| $SP_a$ | Spatial Attention(K:3, R:16) |
| **Layer** | $E_r$ : **Encoder(reference input)** |
| $E_{r1}$ | Conv(In:1, O:64, K:7, P:1, S:1), BNorm, ReLU |
| $E_{r2}$ | Conv(In:64, O:128, K:3, P:1, S:1), BNorm, ReLU |
| $E_{r3}$ | Conv(In:128, O:256, K:3, P:1, S:1), BNorm, ReLU |
| $E_{r4}$ | Conv(In:256, O:512, K:3, P:1, S:1), BNorm, ReLU |
| $CH_a$ | Channel Attention(K:3, R:16) |
| **Layer** | **Resblocks** |
| L1,2,3,4 | Conv(In:1024, O:512, K:3, P:1, S:1), BNorm, ReLU |
| **Layer** | **Decoder** |
| Decoder1 | Conv(In:512, O:256, K:4, P:2, S:1), BNorm, ReLU |
| Decoder2 | Conv(In:256, O:128, K:7, P:3, S:1), BNorm, ReLU |
| Decoder3 | Conv(In:128, O:64, K:7, P:3, S:1) |
| Decoder4 | Conv(In:64, O:1, K:7, P:3, S:1) |
| Function | Hyperbolic tangent(L3) |
| **Color image domain Generator $G_c$** | |
| **Layer** | $E_r$ : **Encoder** |
| Encoder1 | Conv(In:1, O:64, K:7, P:1, S:1), BNorm, ReLU |
| Encoder2 | Conv(In:64, O:128, K:3, P:1, S:1), BNorm, ReLU |
| Encoder3 | Conv(In:128, O:256, K:3, P:1, S:1), BNorm, ReLU |
| Encoder4 | Conv(In:256, O:512, K:3, P:1, S:1), BNorm, ReLU |
| **Layer** | **Resblocks** |
| L1,2,3,4 | Conv(In:512, O:512, K:3, P:1, S:1), BNorm, ReLU |
| **Layer** | **Decoder** |
| Decoder1 | Conv(In:512, O:256, K:4, P:2, S:1), BNorm, ReLU |
| Decoder2 | Conv(In:256, O:128, K:7, P:3, S:1), BNorm, ReLU |
| Decoder3 | Conv(In:128, O:64, K:7, P:3, S:1) |
| Decoder4 | Conv(In:64, O:3, K:7, P:3, S:1) |
| Function | Hyperbolic tangent(L3) |

Table 2: Three different input sketches (Anchor, Positive, Negative) are fed into the convolution layers which share the same weights.

| Sketch style loss base model $f$ | |
|---|---|
| **Layer** | $L1 - 4$: **Encoder(Anchor, Positive, Negative)** |
| $L1$ | Conv(In:1, O:64, K:7, P:3, S:1), BNorm, ReLU |
| $L2$ | Conv(In:64, O:128, K:4, P:1, S:2), BNorm, ReLU |
| $L3$ | Conv(In:128, O:256, K:4, P:1, S:2), BNorm, ReLU |
| *Pooling* | Flatten(AdaptiveAvgPool2d(1x1)) |
| $L4$ | Linear(In:256, O:128) |

To check how well the contrastive network pre-trained model determines the style of sketches, we apply sketches from our 4SKST dataset. We extract the features of sketches using the pre-trained model and cluster the features with T-SNE [Van der Maaten and Hinton 2008] to visualize how well different sketch styles are distributed separately. See Figure 1 for the results. The figure shows that four different styles are clustered in different locations relatively well, although style 1 and style 4 are somewhat intertwined because they share visually similar sketch styles. See Figures 5 and 6 for the sketches from 4SKST dataset and check how they are visually similar as the T-SNE distribution indicates.

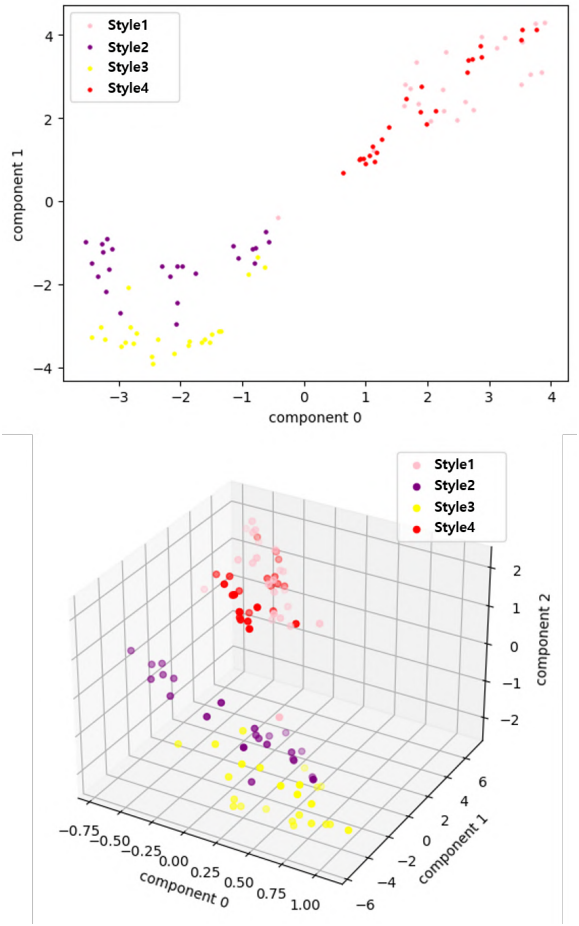

Figure 1: We extract the features of sketches from 4SKST dataset using the pre-trained contastive model and clustered the features by T-SNE method. For visualization, we align the extracted features in 2-dimensional and 3-dimensional spaces.

## 3 DATASET

As described in the main paper, we apply K-means clustering [Lloyd 1982] with K=4 to define four major sketch drawing styles from the training sketch dataset. To cluster the images, we use the image features extracted from the first fully connected layer ($7 \times 7 \times 512$) of the pre-trained VGG16 model [Simonyan and Zisserman 2014]. Before clustering the sketches to find four major styles, we first removed incorrectly tagged images by applying K-means clustering to the retrieved images from safebooru to make an improved dataset. Many images which are not suitable as a sketch data are culled out by the clustering of K=10, such as color images, low quality sketches, semi-colored line art, images with transparent background, and etc. See Figure 2 for the clusters of improper images. We repeated this clustering three times to completely remove improper images from the dataset. We then apply K-means clustering again to the improved dataset with K=4. The images of four different styles are clustered to similar styles and our artist imitated these styles to draw authentic sketches corresponding to the given color image. The artist drew all of the sketches with a pen tablet device and drawing software [CELSYS 2012]. See Figure 3 for a visualization of clustering result and Figures 5 and 6 for dataset examples.

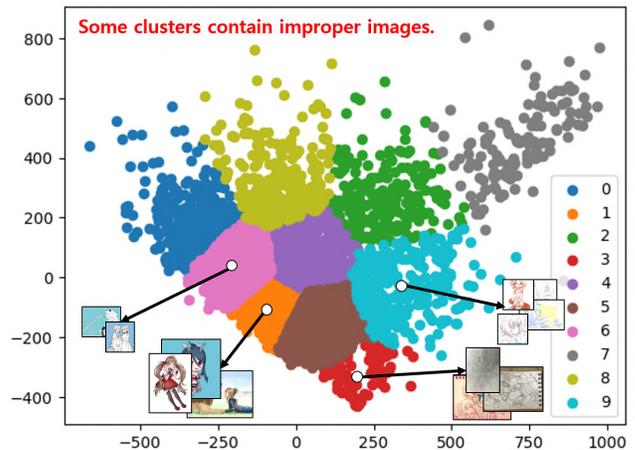

Figure 2: K-means clustering applied to line art tagged images from Safebooru [DanbooruCommunity 2021]. Improper images are clustered to have same labels. (Example images are included for an illustration purpose) © Chobi, BioTroy, AkaneNagano

## 4 BASELINE COMPARISON DETAILS

As mentioned in the main paper, baseline methods are trained with the same dataset as ours except Ref2sketch [Ashtari et al. 2022]. The settings are based on their official code and information from their respective papers. MUNIT [Huang et al. 2018], Park et al. [2020], Council-GAN [Nizan and Tal 2020], and IrwGAN [Xie et al. 2021] are trained with a total of 100k iterations with the batch size of 4. Chan et al. [2022] is trained with 100 epochs total with the batch size of 4. The ablation study and all output images from the baseline methods are evaluated with three different metrics [Heusel et al. 2017; Wang et al. 2004; Zhang et al. 2018a] with the image resolution of $512 \times 512$.

## 5 EFFECTS OF SPATIAL AND CHANNEL ATTENTION

As described in the main paper, our model utilizes spatial and channel attention individually, and concatenates the resulting features to each ResBlock layer after normalizing them using the ADAIN method. It is well-known to emphasize shape features with spatial attention and style features with channel attention individually to enhance the quality of the generated output images. Examples that use the similar approach can be found in the work of Deng et al. [2020] and Fu et al. [2019].

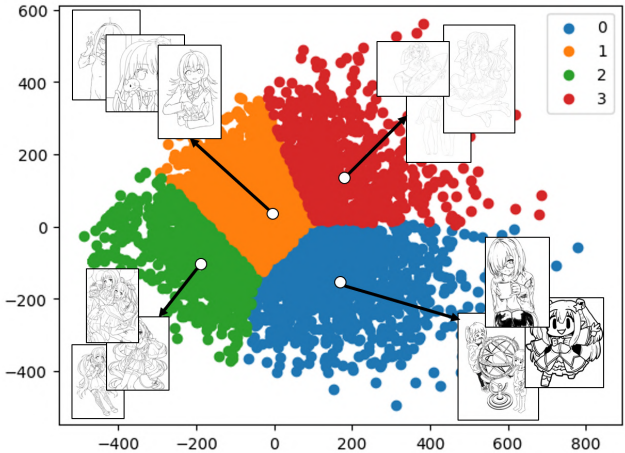

Figure 3: Examples of four major sketch styles identified by K-means clustering. (Example images are included for an illustration purpose) © Chobi, Comete_atr, Ayul, AonoriwaKame

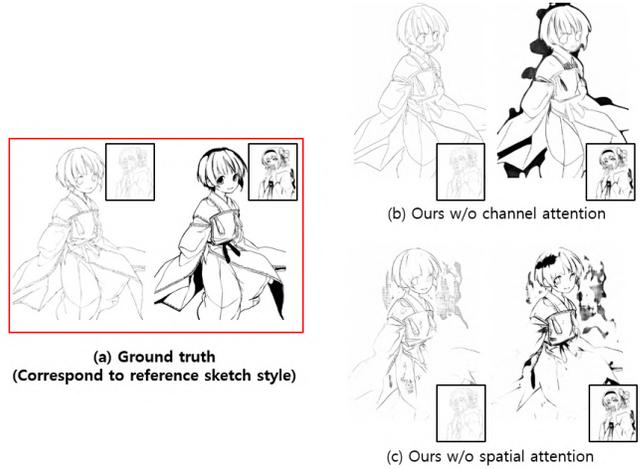

Figure 4: Example results from the ablation study on each attention function. © 4SKST

Table 3: Quantitative comparison of our method without spatial or channel attention concatenation and Ours+$\mathcal{L}_{clip}$ method.

| Method | PSNR↑ | FID↓ | LPIPS↓ |
|---|---|---|---|
| Ours w/o channel attention | 34.30 | 109.48 | 0.2131 |
| Ours w/o spatial attention | 34.90 | 147.52 | 0.2790 |
| Ours | **35.58** | **82.18** | 0.1271 |
| Ours+$\mathcal{L}_{clip}$ | 35.32 | 89.32 | **0.1190** |

To prove the benefit of using both types of attention, we conducted an ablation study by removing each attention function before normalizing the features. For example, if spatial attention is removed, the $ADAIN_a$ in the network normalizes the reference sketch input image features emphasized by channel attention and the color input image features without spatial attention. We reported a quantitative comparison results of this ablation study in Table 3. If one of these attention functions is removed, the performance became poorer. For qualitative comparison examples, please refer to Figure 4.

Comparison of (b) and (c) to ground truth sketches (a), which correspond to the input reference sketch style, shows that the results produced without channel attention (b) have the fully generated shape of the character, but details such as the eyes and shading are incorrect. Similarly, the results produced without spatial attention (c) have some details similar to the ground truth, but the overall shape and lines are not fully constructed.

## 6 ADDING CLIP BASED SEMANTIC LOSS

Similar to Chan et al. [2022], we conducted an experiment by additionally implementing the semantic loss that utilizes the CLIP [Radford et al. 2021] method. The shared visual-text embedding CLIP extracts semantic information from both color and sketch images. The semantic loss function minimizes the distance between the CLIP embeddings of the color input image and sketch output. The implementation detail is the same as the official code from Chan et al. [2022]. The loss is expressed as follows:

$$\mathcal{L}_{clip} = ||CLIP(C_i) - CLIP(O)||] \quad (2)$$

$C_i$ denotes the color input image and $O$ denotes the output sketch. We train the model that has this semantic loss added to our method. The training details are same as our method described in the main paper. A quantitative comparison between methods Ours and Ours+$\mathcal{L}_{clip}$ is reported in Table 3. Ours achieves better scores in PSNR and FID while Ours+$\mathcal{L}_{clip}$ achieves better scores in LPIPS. Moreover, the visual comparison of outputs does not clearly indicate which method is better in all examples. Meanwhile, adding $\mathcal{L}_{clip}$ requires twice more computation times when training the model (Ours: 0.47 secs per iteration, Ours+$\mathcal{L}_{clip}$: 1.16 secs per iteration). Therefore, we decide not to include $\mathcal{L}_{clip}$ in our method. See Figure 7 for qualitative examples.

## 7 PERCEPTUAL STUDY DETAILS

The question of *"Which sketch style looks more similar to the target sketch?"* was was asked to 200 participants for a total of 20 comparisons. See examples of the comparisons in Figures 8,9,10,11 and 12.

## 8 IMPROVING AUTO-COLORIZATION DETAILS

A model for auto-colorizing sketches requires a large amount of data for sketches and their corresponding colorized images. Therefore,

existing methods utilize sketch-extraction techniques to gather synthetic sketch images from colorized images to train their models. See Table 4 for the training dataset information of related methods. However as mentioned in the main paper, using synthetic sketches of only a single style for training can cause over-fitting to the model due to the scarcity of various sketch styles information.

For the experiments performed in this section, we used Ci et al. [2018] as the auto-colorization model base. The code was implemented based on the original paper and contains necessary functions such as the color-guided concatenation and optimizations to enhance the quality. To prove the benefit of the multi-modal sketch extraction method, we chose baseline sketch extraction methods that are used to generate the dataset for training existing auto-colorzation models. See Table 4 for sketch extraction methods used for training existing methods.

We also trained the model with the combination of all styles of extracted sketches except ours to show the superiority of our method in extracting multi-styles sketches. To train the model, 1,500 color images were retrieved from safebooru [DanbooruCommunity 2021] and the sketches were extracted by each method. The total training epochs for each dataset were 800. In contrast, the models trained with the combined [Canny 1986; lllyasviel 2017; Simo-Serra et al. 2016; Winnemöller 2011] dataset and the dataset generated by our method were trained for only 200 epochs. This is because each baseline method extracted sketches of only one style and has only 1,500 pairs of color and sketch images. Our method extracted sketches of four different styles from the 1,500 color images, making a total 6,000 of pairs of color and sketch images. The combined dataset also has 6,000 pairs of images. To test the model, we used 4 SKST dataset that consists of four different styles of authentic sketches and paired color images. The quantitative results from the evaluation metrics are described in Table 4 of the main paper. See Figures 13 and 14 for examples of the auto-colorized output trained with the different datasets.

**Table 4: A list of sketch auto-colorization methods. The training procedures utilized different sketch-extraction methods to create synthetic training data.**

| Methods | Size of training dataset | Dataset-methods |
|---|---|---|
| Auto-painter [Liu et al. 2018] | Over 60K | XDoG |
| Thasarathan et al. [Thasarathan and Ebrahimi 2019] | 100K for each weight | Canny |
| Tag2Pix [Kim et al. 2019] | Over 54K | Sketch Keras + XDoG |
| Style2Paints [Zhang et al. 2018b] | Over 3.33m | Sketch Keras |
| AlacGAN [Ci et al. 2018] | Over 22K | XDoG |
| Lee et al. [2020] | Different by domain | XDoG |
| Yuan and Simo-Serra [2021] | Over 1,229K | Sketch Keras + XDoG +Simo-Serra et al. [2016] |

## 9 SKETCH STYLE TRANSFER

As outlined in the main paper, our approach can transfer the style of the input sketch image. This style transfer approach can also be utilized to simplify digitized rough sketches. To show some examples of this task, we included the sketches simplified by an existing sketch simplification method [Simo-Serra et al. 2018]. Simo-Serra et al. [2018] is a well known semi-unsupervised sketch simplification method based on deep-learning. To simplify rough sketches using Simo-Serra et al. [2018] we utilized the pre-trained weights from the official page of paper. Unfortunately, we found that Simo-Serra et al. [2018] is trained with a different dataset and cannot stylize the output sketches by imitating the reference input. Therefore this qualitative comparison should be used only for an illustration purpose of the sketch style transfer applied to the sketch simplification task. See Figure 15 for the examples.

## 10 QUALITATIVE COMPARISON TO REF2SKETCH

Although our method outperforms the baselines in quantitative evaluations, one might think that the results may be different if the test was performed on different images from 4SKST dataset. Therefore, we included more qualitative examples that compare our method with Ref2sketch. We chose Ref2sketch for this additional comparison because Ref2Sketch achieved mostly better scores than other baselines which can accept the reference input to imitate the style. In the qualitative comparison, Ref2sketch produced as fine quality as our method in anime style images. However, our method produced clearly superior results to Ref2sketch on photo images. This is noteworthy because our method was trained only with anime style images from Danbooru [DanbooruCommunity 2021]. See Figures 16,17,18 and 19 for examples.

## 11 ADDITIONAL EXAMPLES

To show that our method can handle images from various domains, we included additional examples that contain content other than a single person. See Figures 20 and 21 for examples.

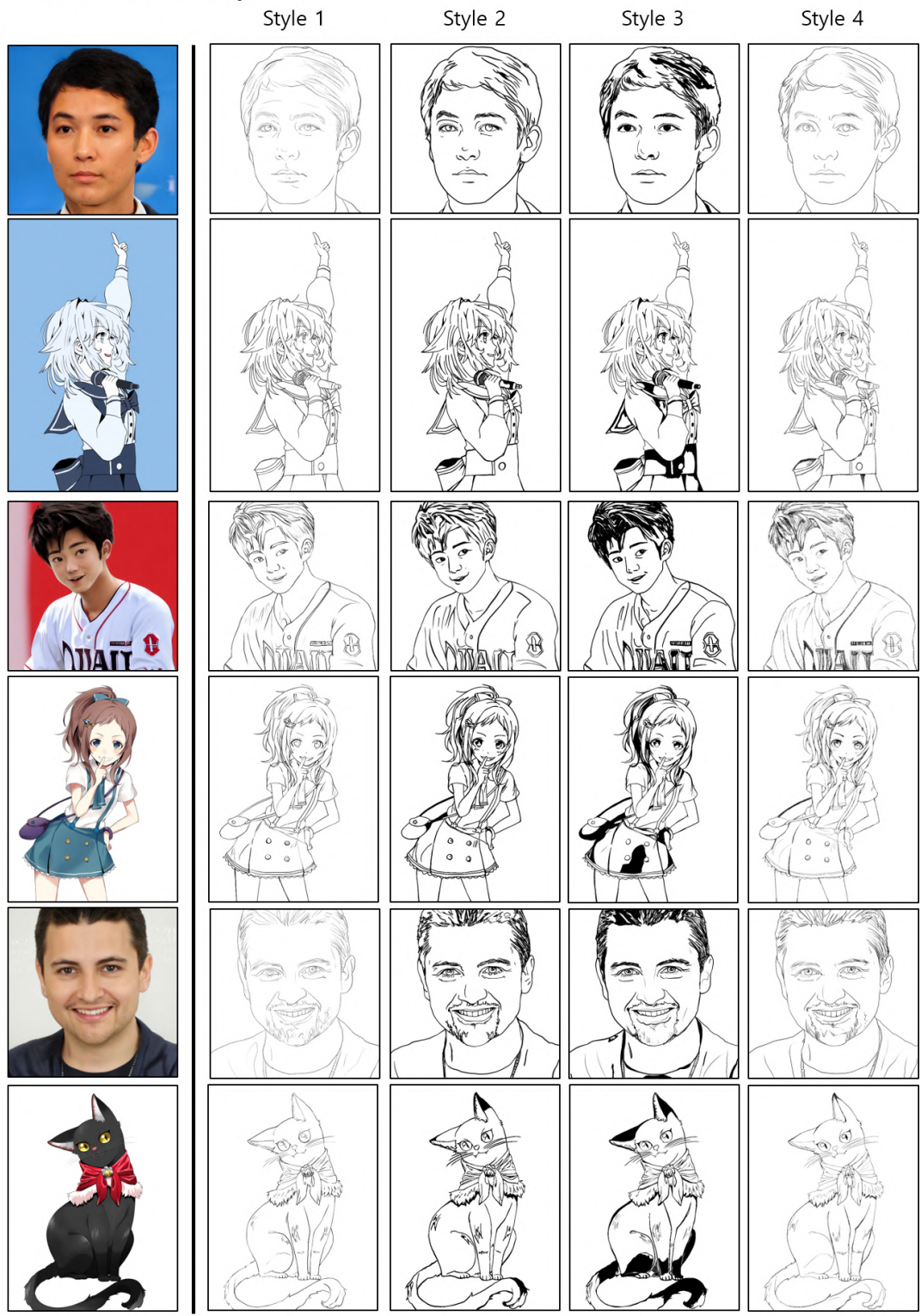

Figure 5: Examples of 4SKST Dataset. It has sketches of one of four different styles paired to each color image. © 4SKST

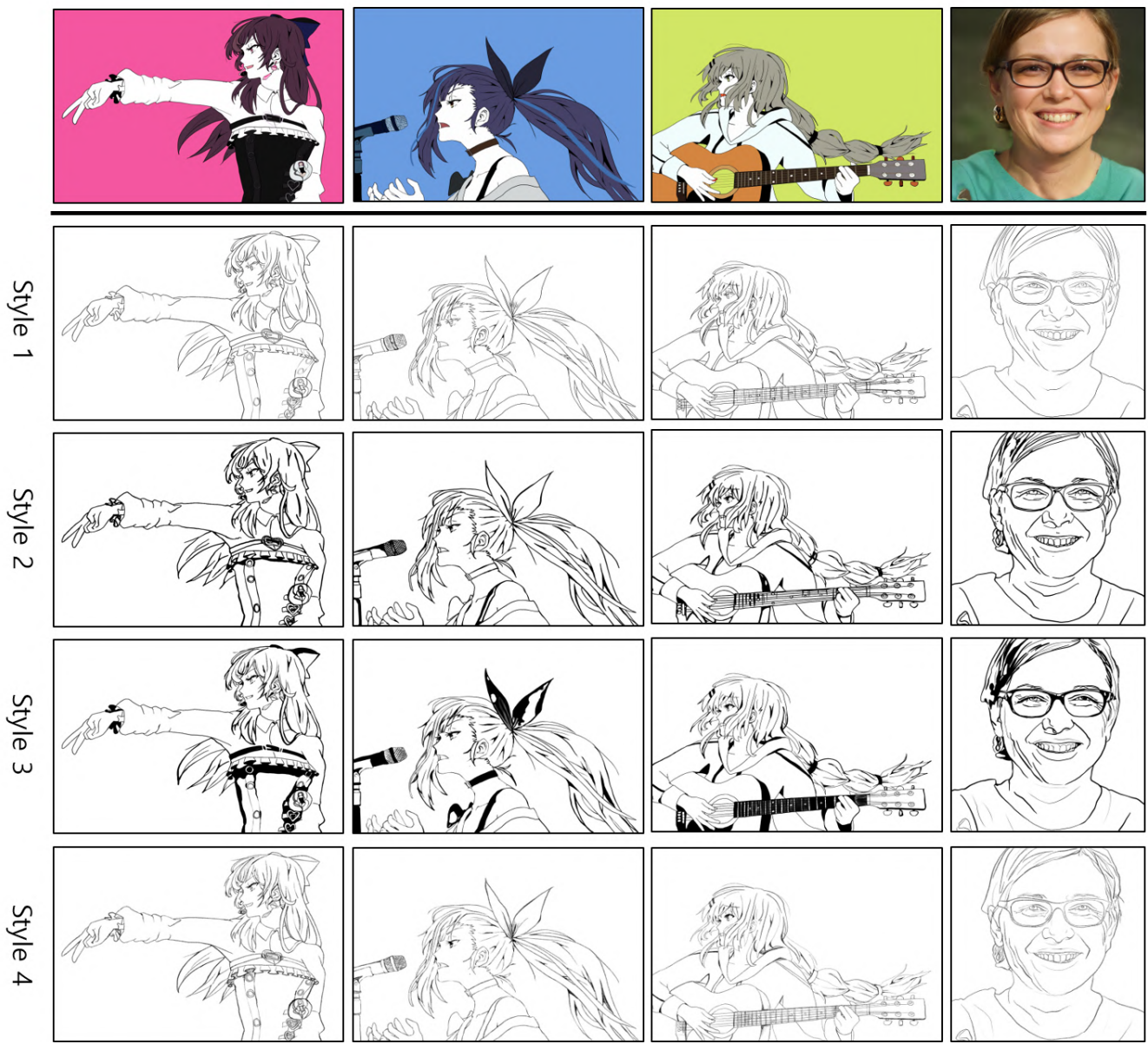

Figure 6: Examples of 4SKST Dataset. It has sketches of one of four different styles paired to each color image. © 4SKST

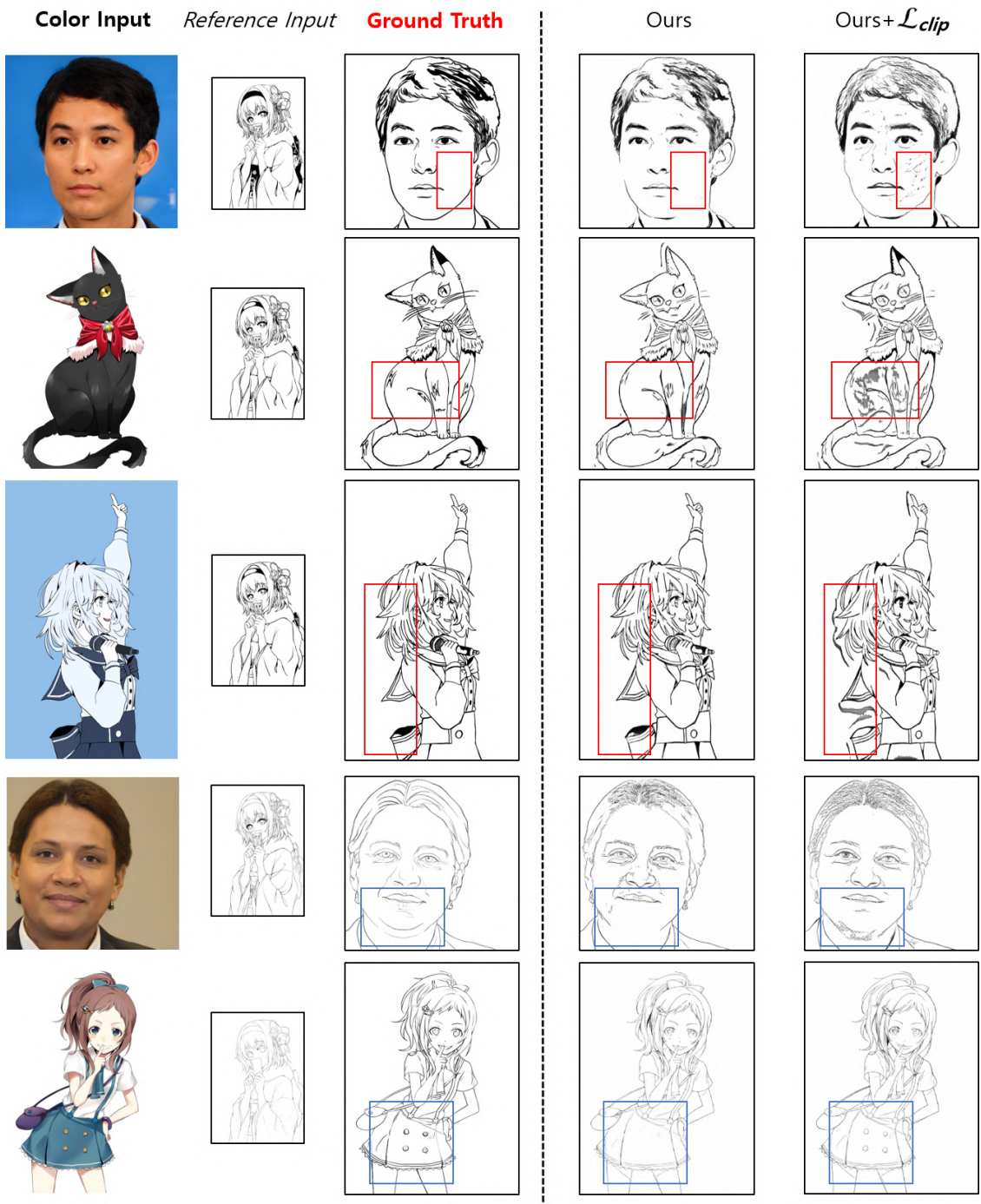

Figure 7: Qualitative comparison between Ours and Ours+$\mathcal{L}_{clip}$. In the rows 1 to 3, some unnecessary artifacts are noticeable on the output from Ours+$\mathcal{L}_{clip}$. Besides, it is apparent that some details are missing on the output from Ours in rows 4 and 5. © 4SKST

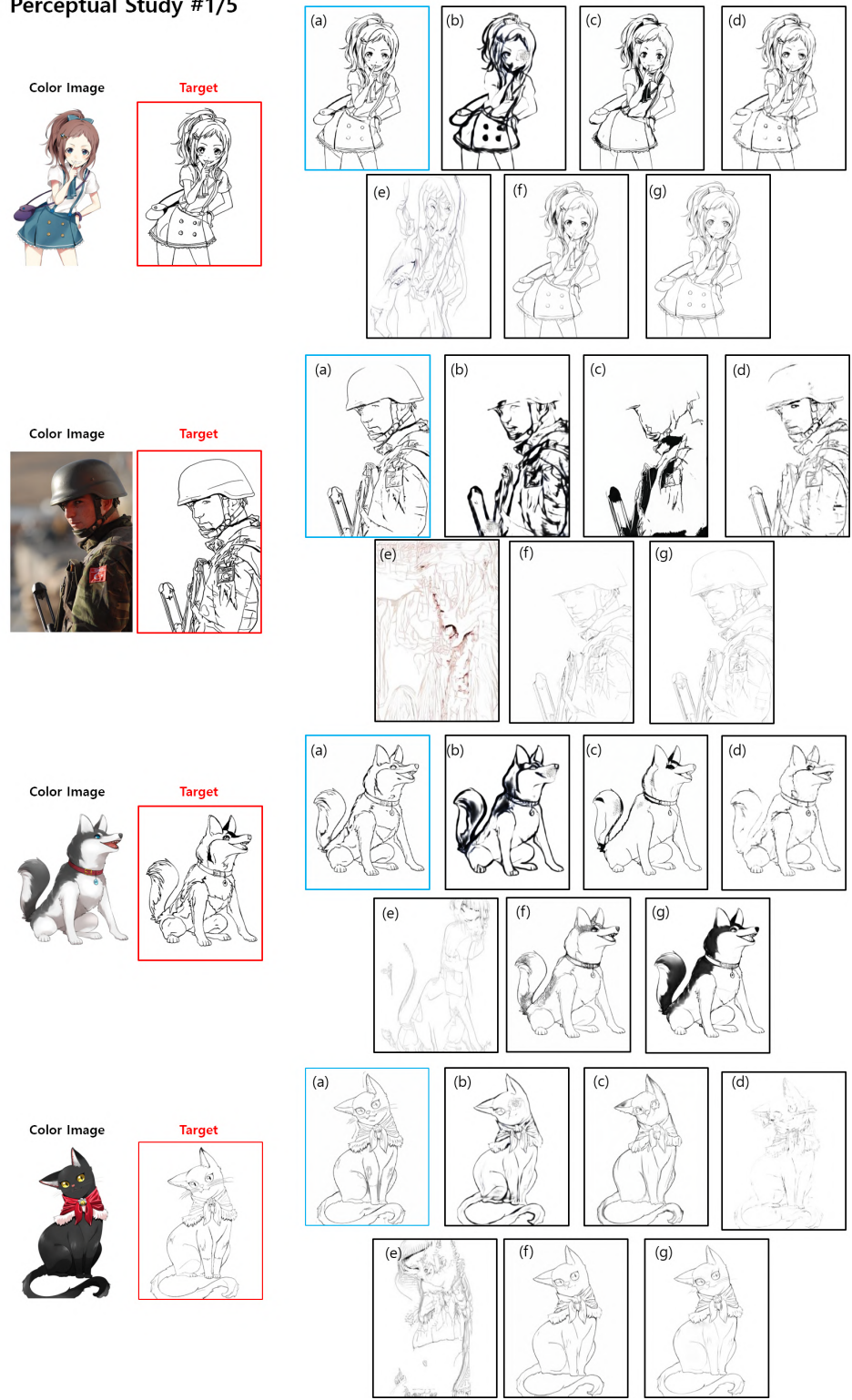

Figure 8: Examples perceptual study. The sketch in a red box is the ground truth and a blue box is the result from our method. The remaining sketches are the outputs from the baseline methods. (a) Ours, (b) MUNIT [Huang et al. 2018], (c)Park et al. [2020], (d)Ref2sketch [Ashtari et al. 2022], (e)Council-GAN [Nizan and Tal 2020], (f)IrwGAN [Xie et al. 2021], (g)Chan et al. [2022]. The color image was not presented while conducting the survey and the display order of the sketches was chosen randomly for each comparison. © 4SKST

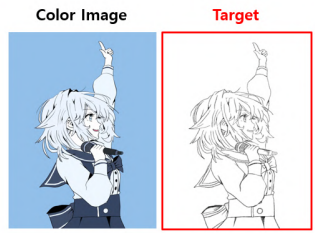
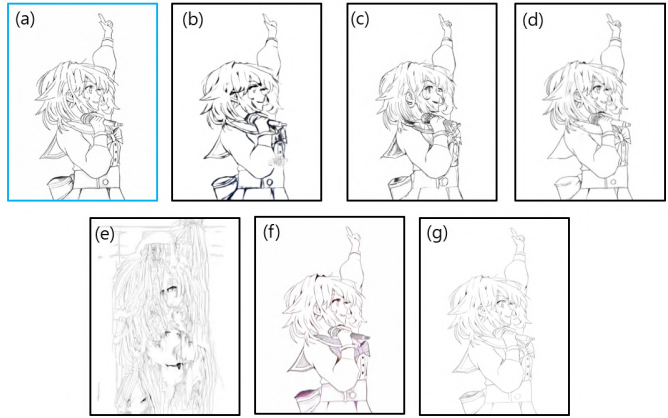
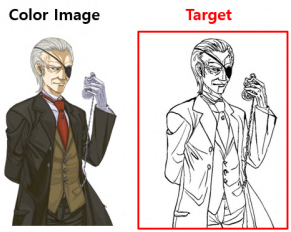
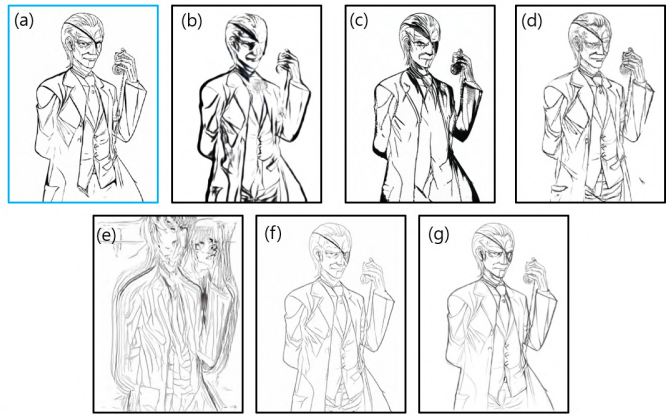
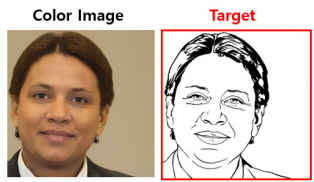
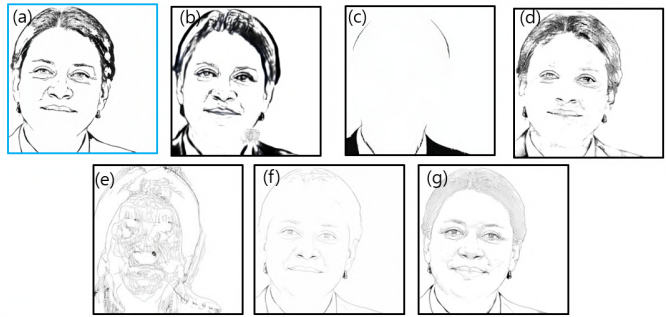
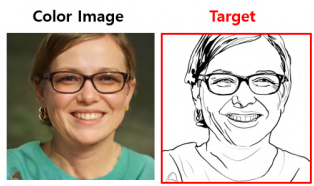
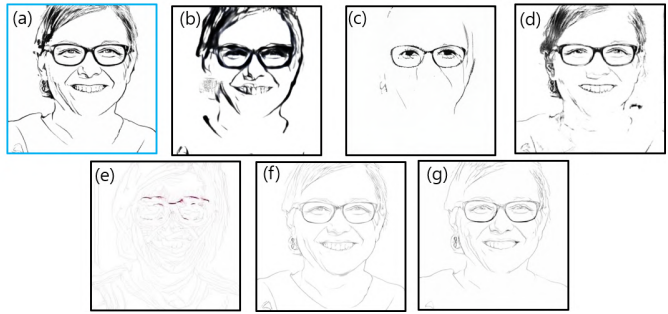

Figure 9: Examples perceptual study. The sketch in a red box is the ground truth and a blue box is the result from our method. The remaining sketches are the outputs from the baseline methods. (a) Ours, (b) MUNIT [Huang et al. 2018], (c)Park et al. [2020], (d)Ref2sketch [Ashtari et al. 2022], (e)Council-GAN [Nizan and Tal 2020], (f)IrwGAN [Xie et al. 2021], (g)Chan et al. [2022]. The color image was not presented while conducting the survey and the display order of the sketches was chosen randomly for each comparison. © 4SKST

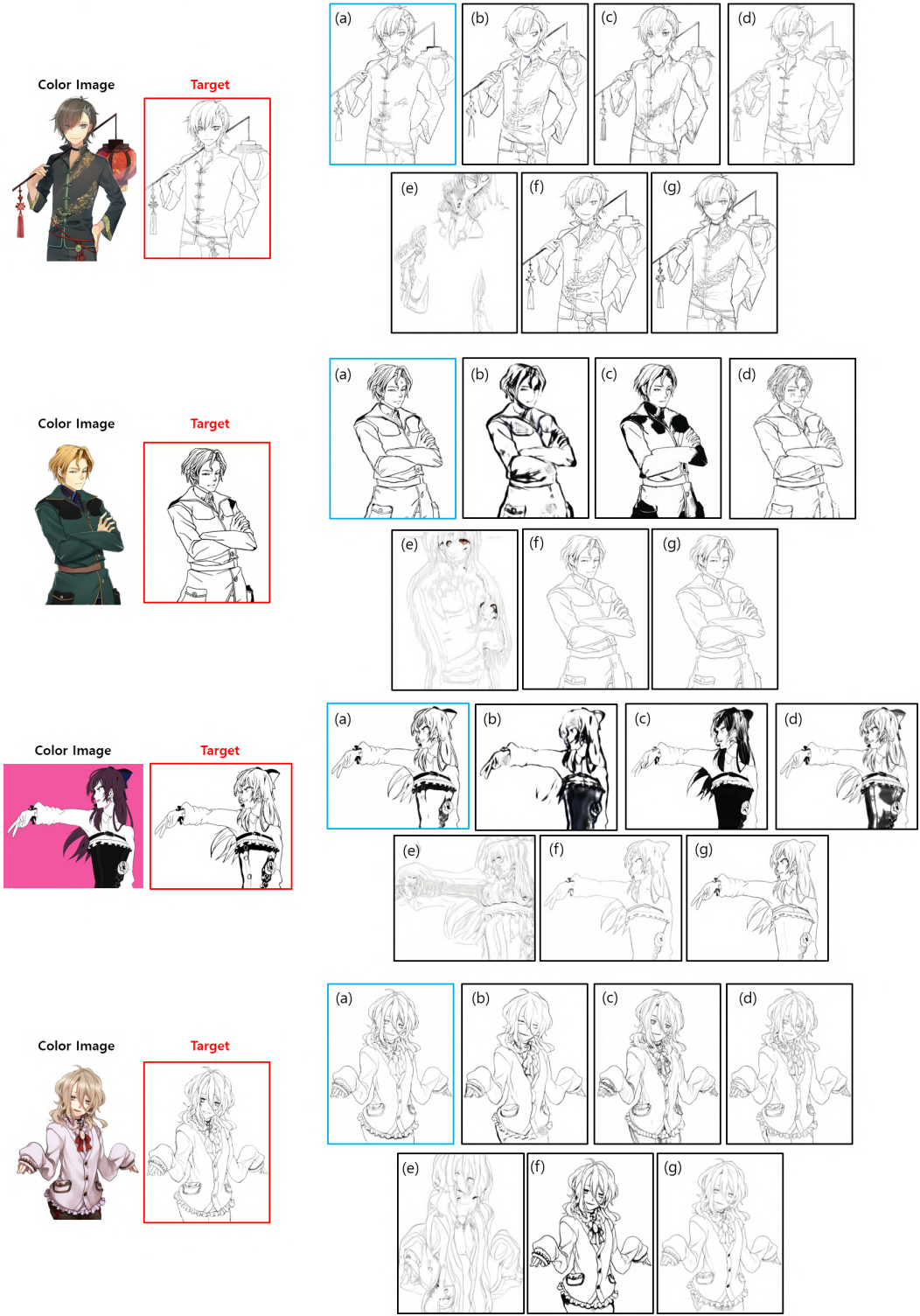

Figure 10: Examples perceptual study. The sketch in a red box is the ground truth and a blue box is the result from our method. The remaining sketches are the outputs from the baseline methods. (a) Ours, (b) MUNIT [Huang et al. 2018], (c)Park et al. [2020], (d)Ref2sketch [Ashtari et al. 2022], (e)Council-GAN [Nizan and Tal 2020], (f)IrwGAN [Xie et al. 2021], (g)Chan et al. [2022]. The color image was not presented while conducting the survey and the display order of the sketches was chosen randomly for each comparison. © 4SKST

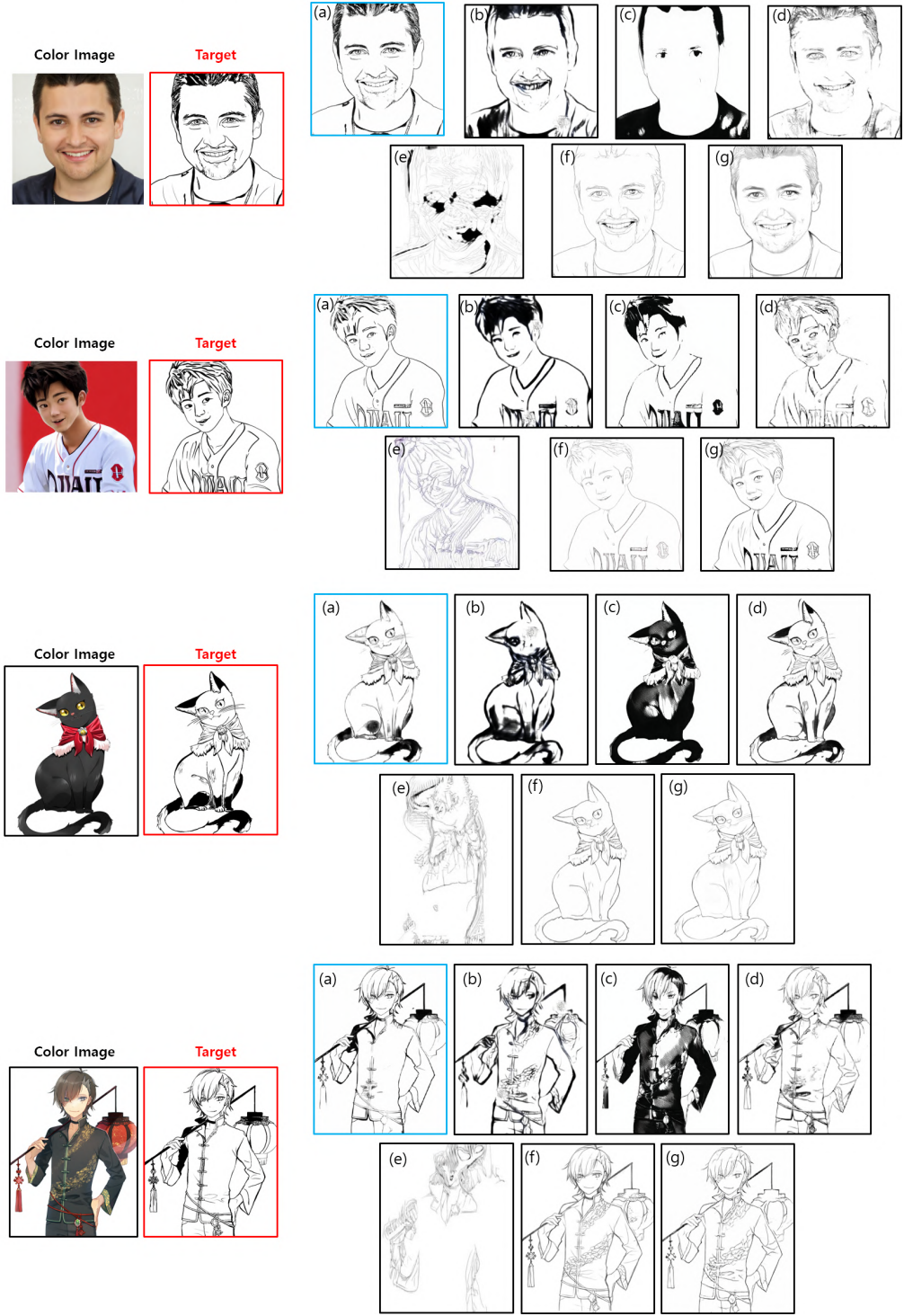

Figure 11: Examples perceptual study. The sketch in a red box is the ground truth and a blue box is the result from our method. The remaining sketches are the outputs from the baseline methods. (a) Ours, (b) MUNIT [Huang et al. 2018], (c)Park et al. [2020], (d)Ref2sketch [Ashtari et al. 2022], (e)Council-GAN [Nizan and Tal 2020], (f)IrwGAN [Xie et al. 2021], (g)Chan et al. [2022]. The color image was not presented while conducting the survey and the display order of the sketches was chosen randomly for each comparison. © 4SKST

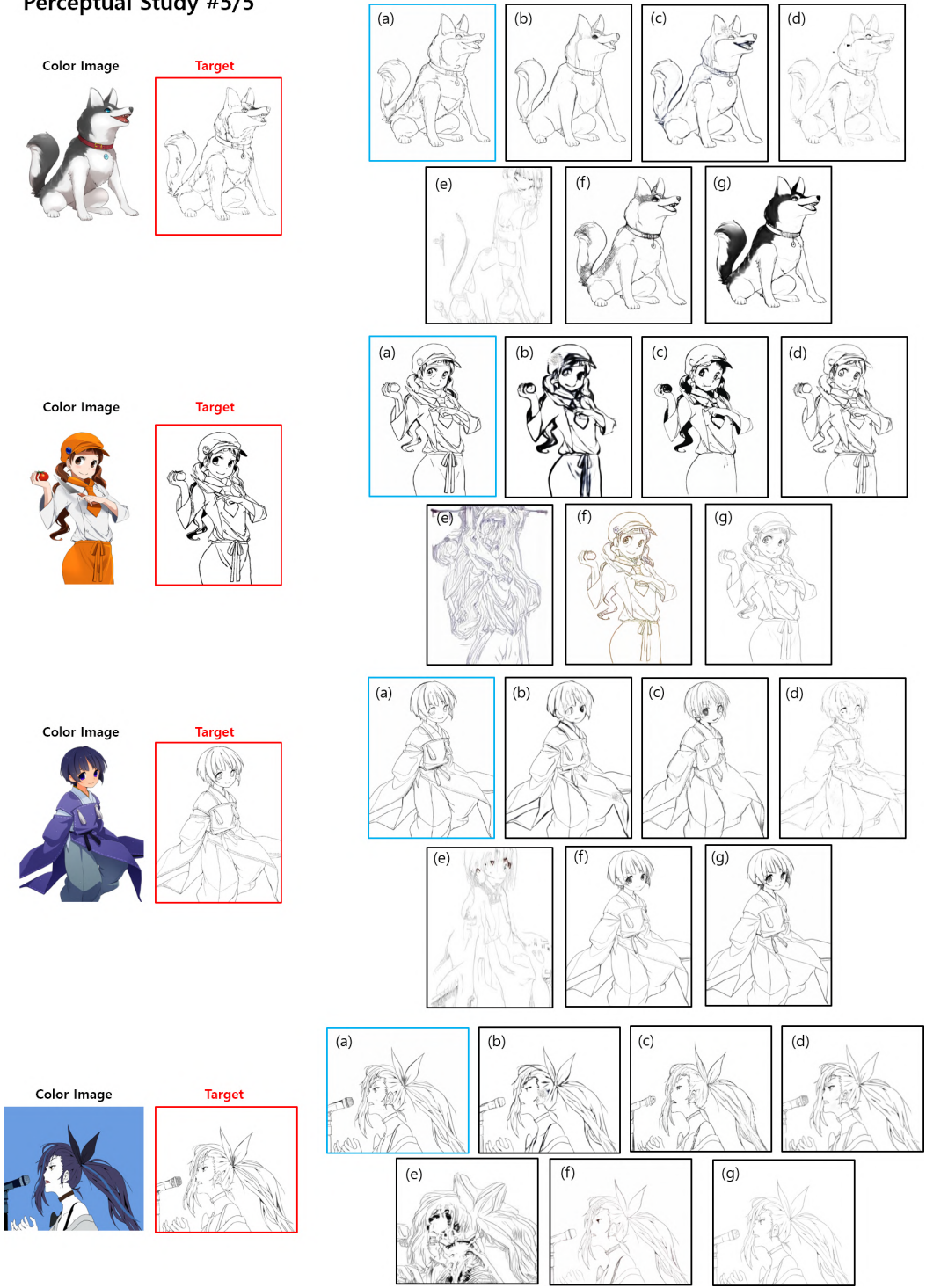

Figure 12: Examples perceptual study. The sketch in a red box is the ground truth and a blue box is the result from our method. The remaining sketches are the outputs from the baseline methods. (a) Ours, (b) MUNIT [Huang et al. 2018], (c)Park et al. [2020], (d)Ref2sketch [Ashtari et al. 2022], (e)Council-GAN [Nizan and Tal 2020], (f)IrwGAN [Xie et al. 2021], (g)Chan et al. [2022]. The color image was not presented while conducting the survey and the display order of the sketches was chosen randomly for each comparison. © 4SKST

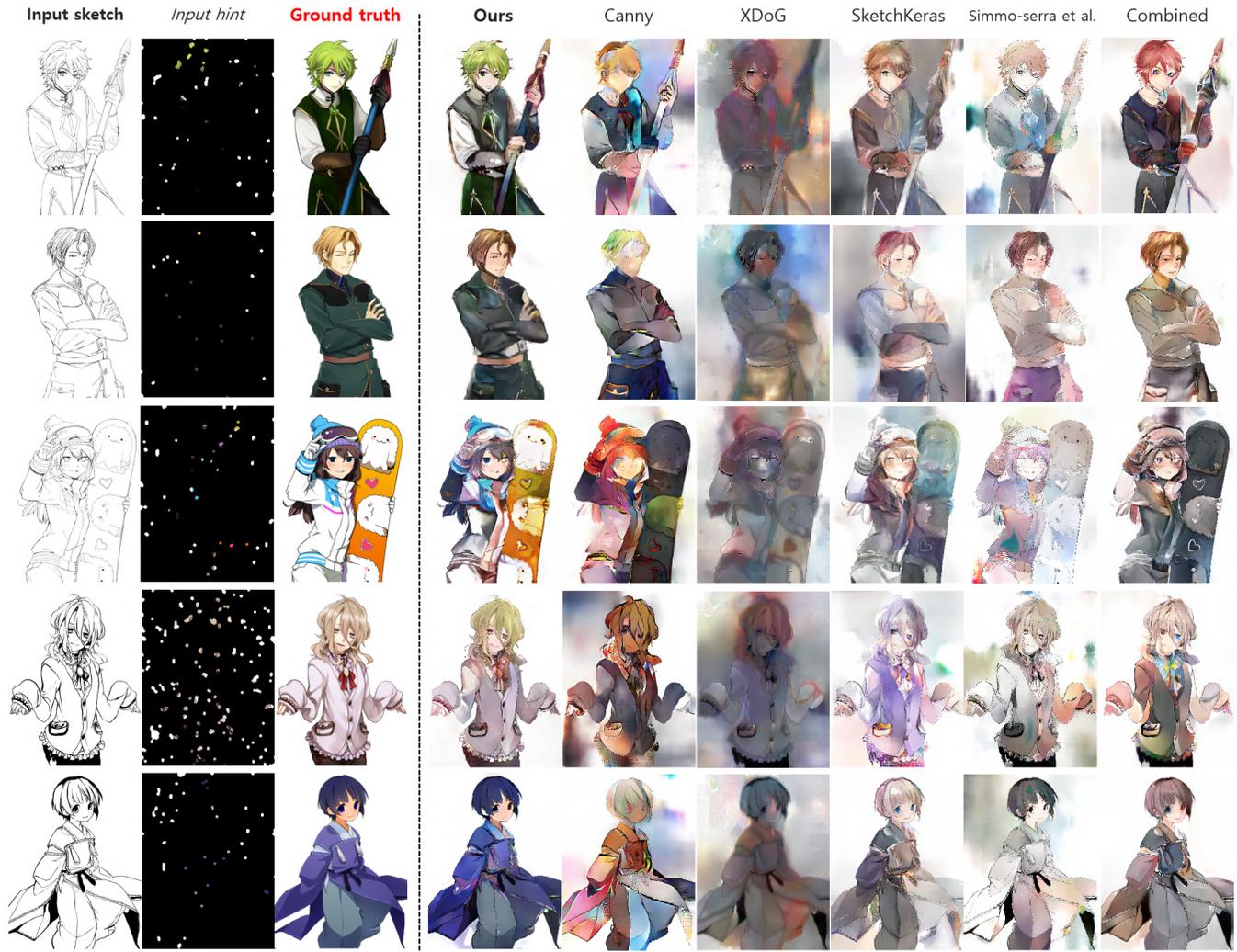

Figure 13: Examples of the output from the auto-colorization method [Ci et al. 2018] trained with different datasets. © 4SKST

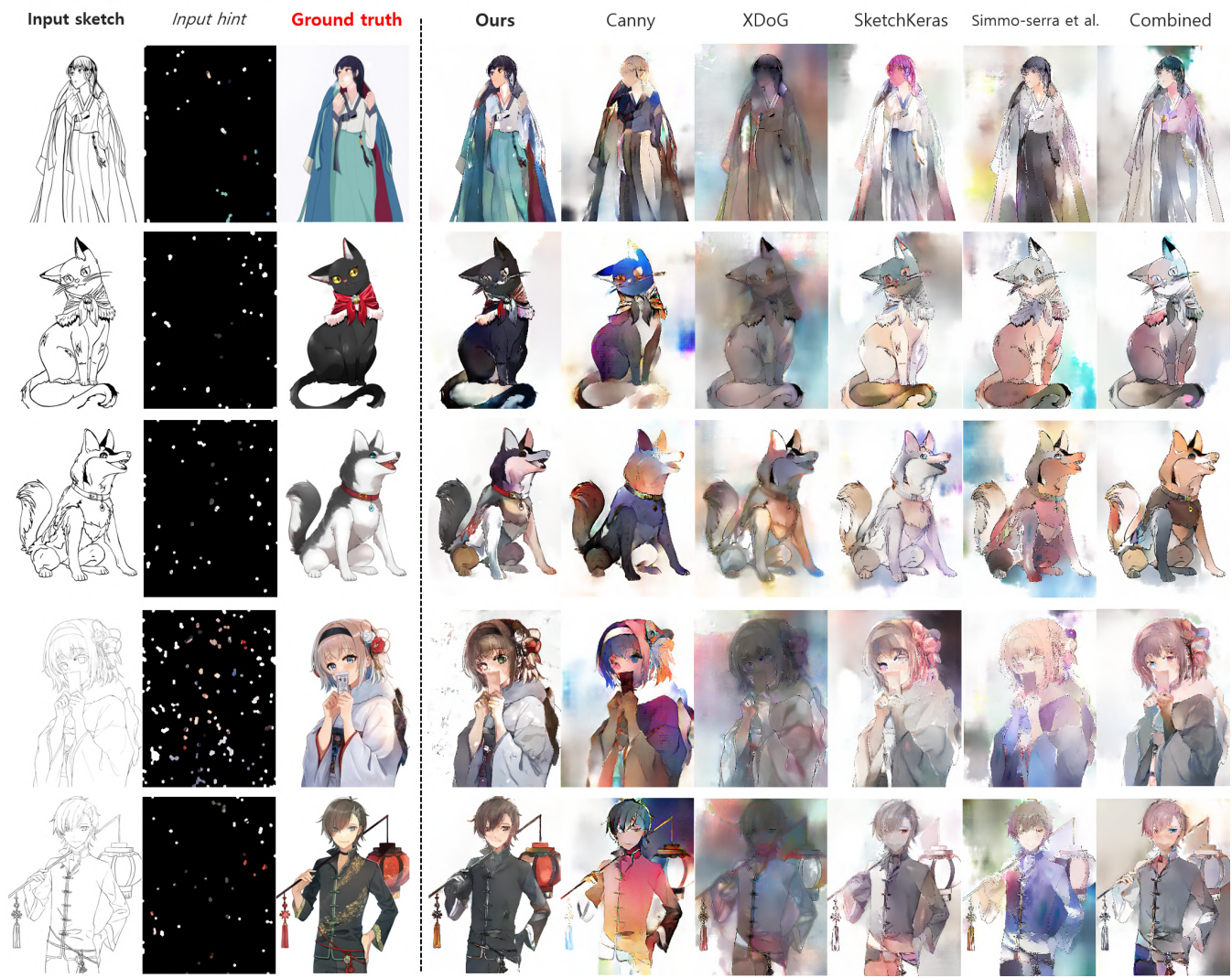

Figure 14: Examples of the output from the auto-colorization method [Ci et al. 2018] trained with different datasets. © 4SKST

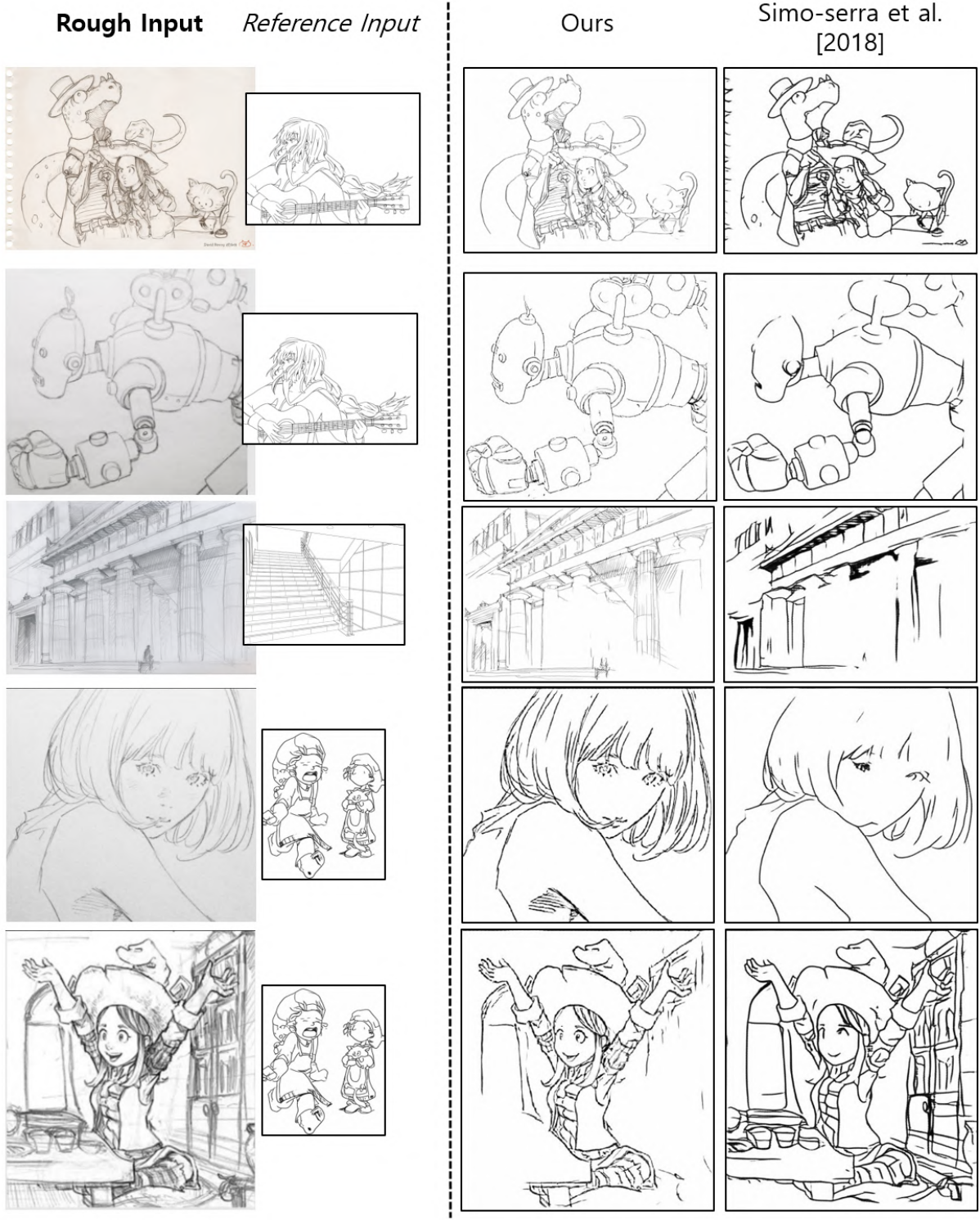

**Figure 15:** Examples of sketch style transfer for simplification produced using our method and Simo-Serra et al. [2018]. © EISAKUSAKU, David Revoy, Yan et al. [2020]

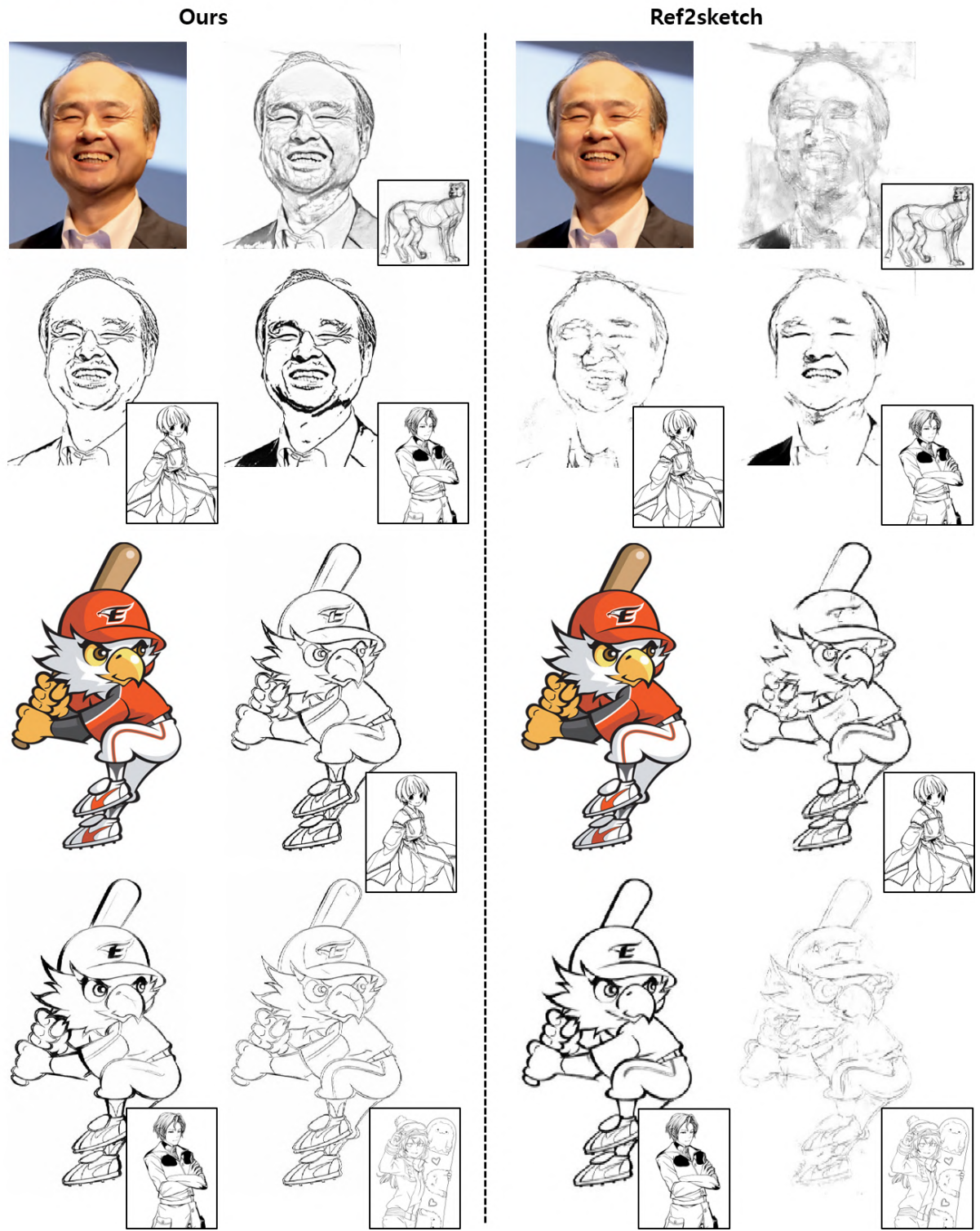

Figure 16: Examples of sketch extraction produced using our method and Ref2sketch[Ashtari et al. 2022]. © KPF, Hanhwa Eagles

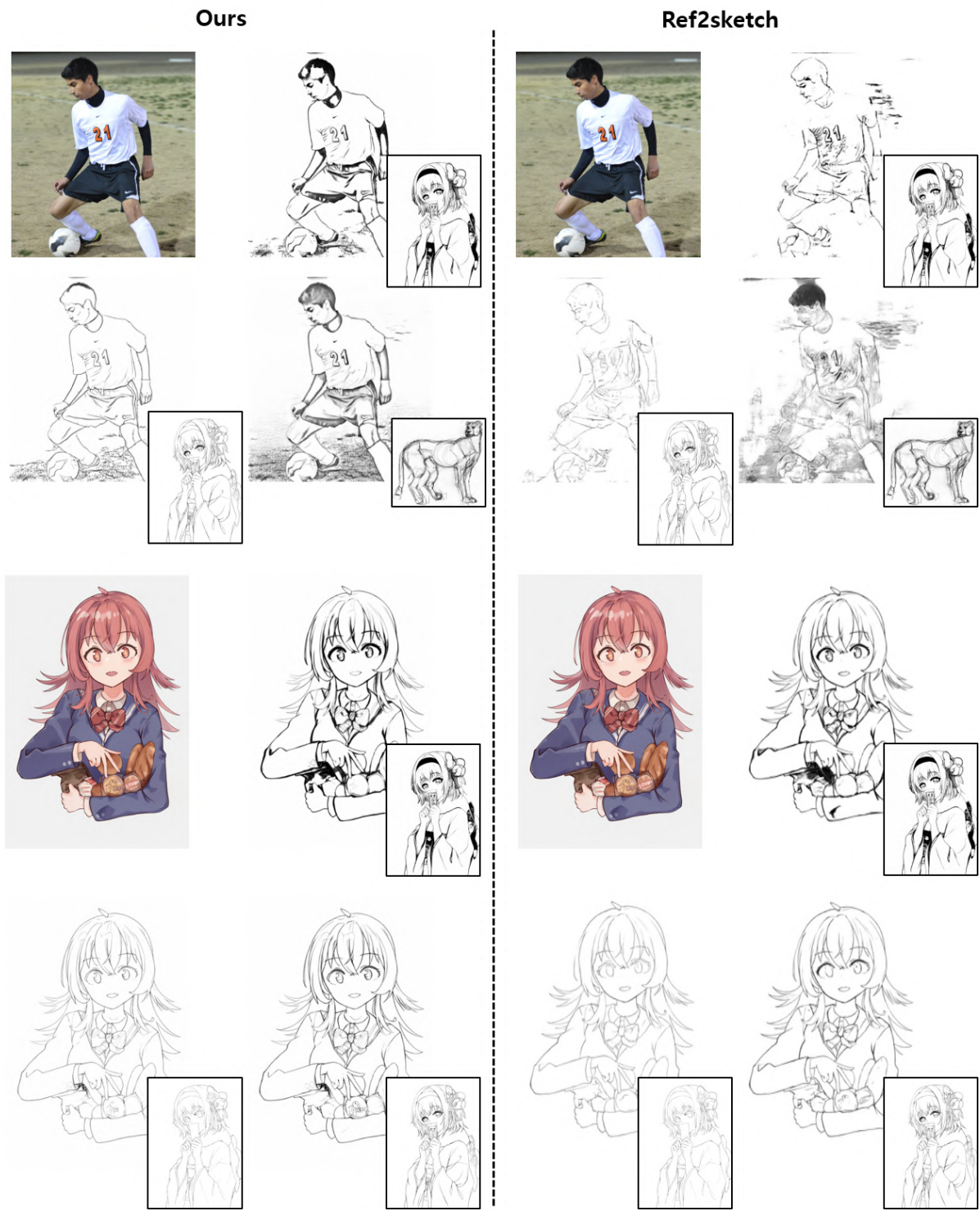

Figure 17: Examples of sketch extraction produced using our method and Ref2sketch[Ashtari et al. 2022]. © selmahighsoccer, Comet_atr

## Qualitative comparison #3/4

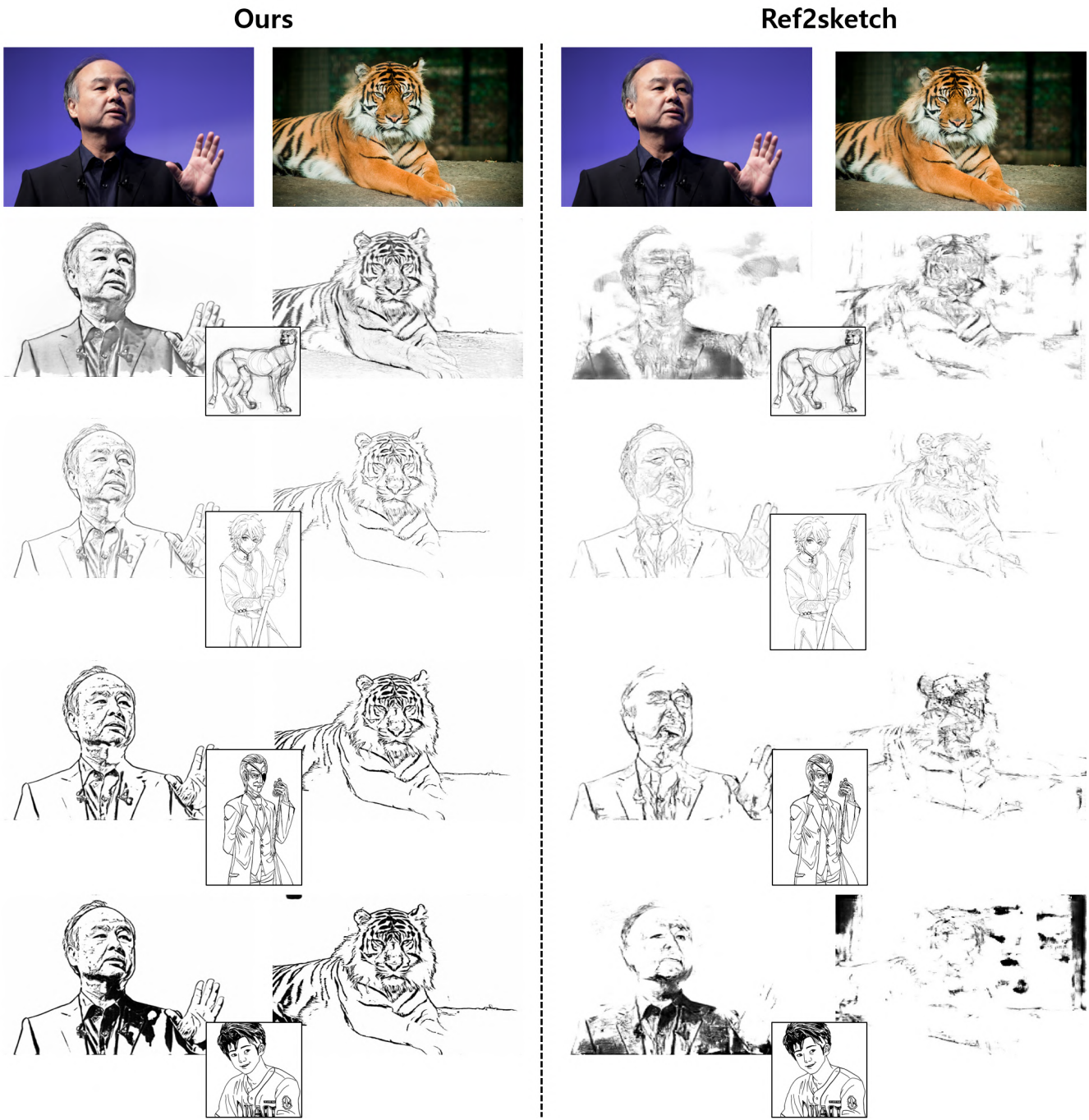

Figure 18: Examples of sketch extraction produced using our method and Ref2sketch[Ashtari et al. 2022]. © KPF, Felinest

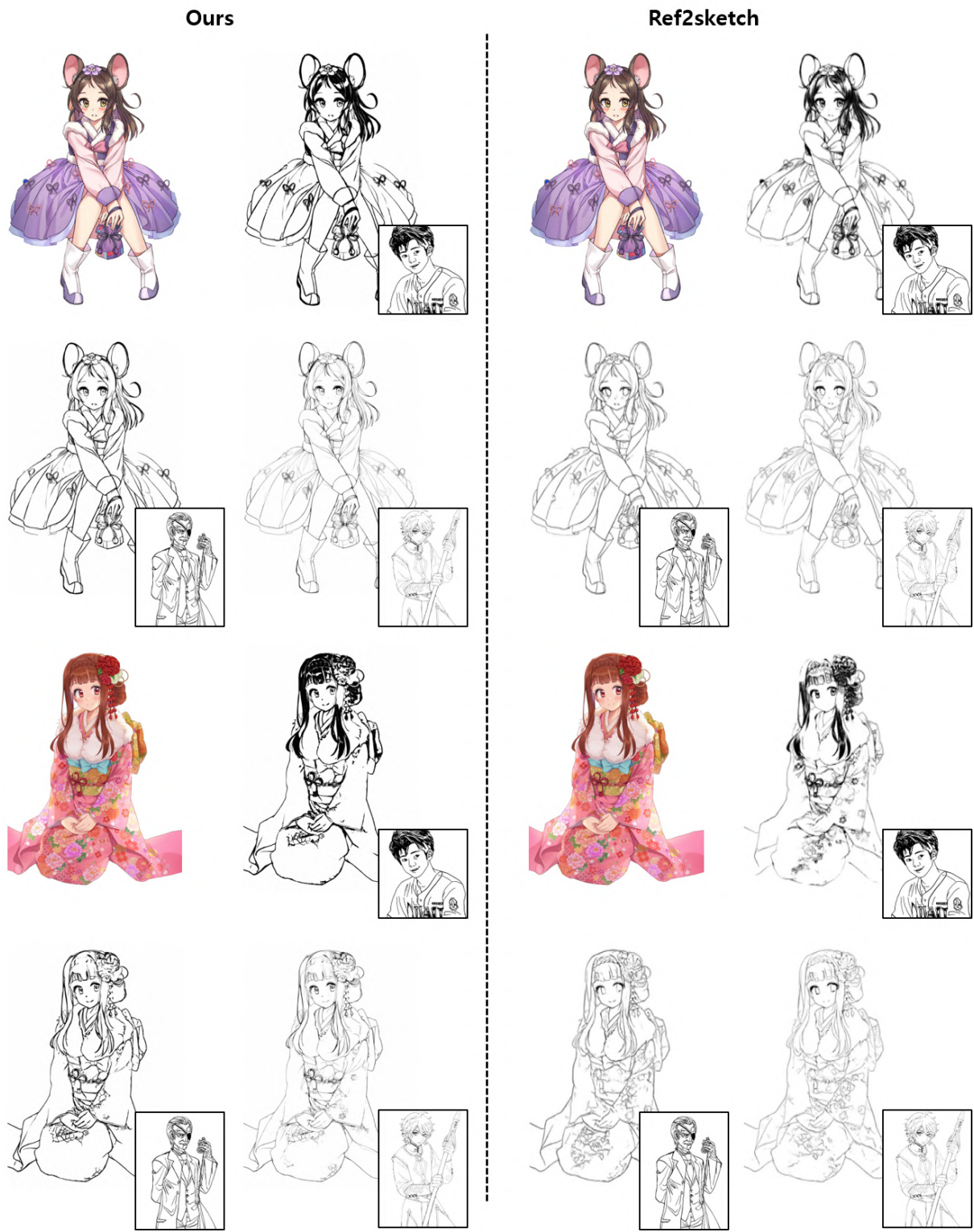

Figure 19: Examples of sketch extraction produced using our method and Ref2sketch[Ashtari et al. 2022]. © Chobi

## More examples of Our method #1/2

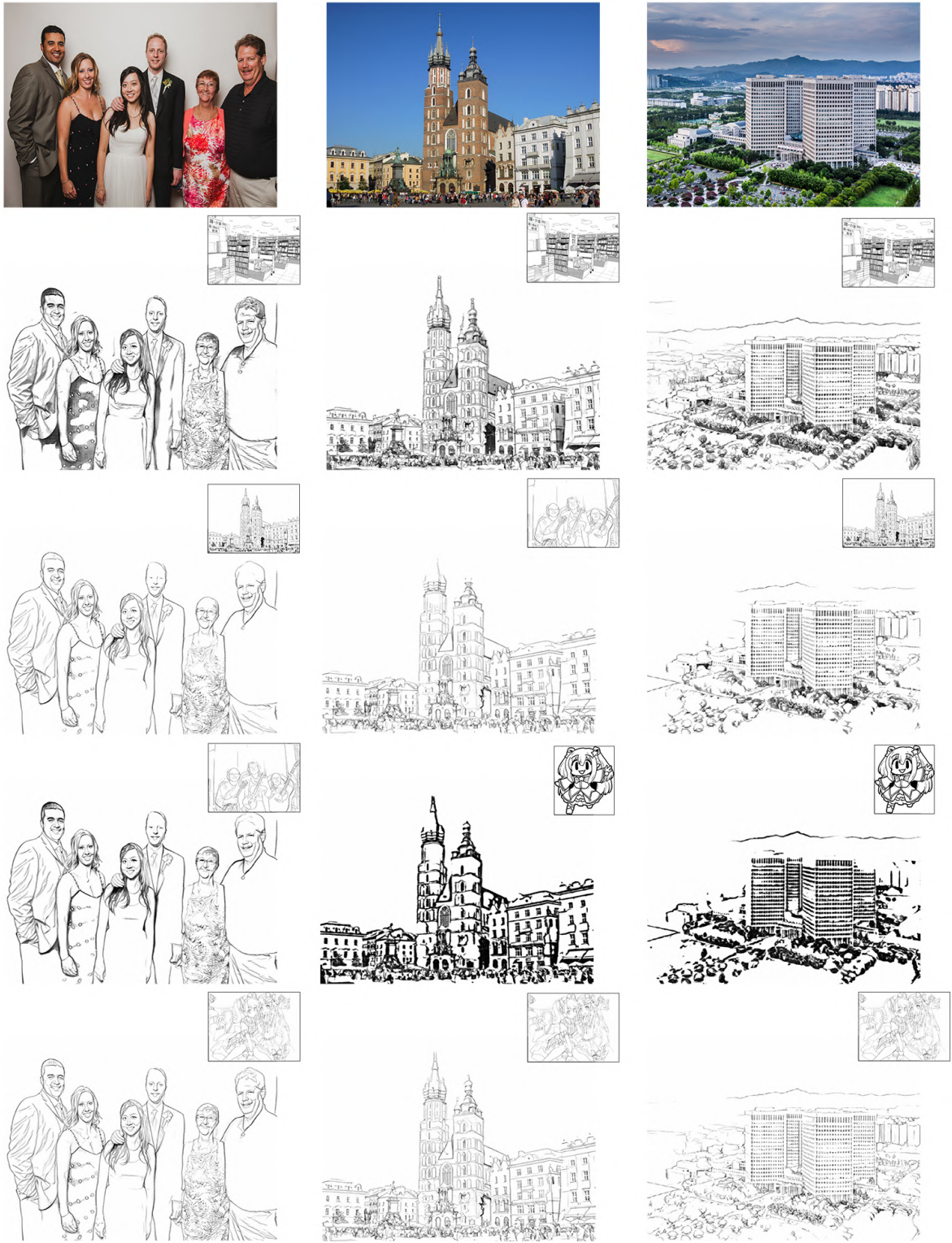

Figure 20: Additional examples of sketch extraction produced using our method. © peteandcharlotte, Güldem Üstün, DDIC

# More examples of Our method #2/2

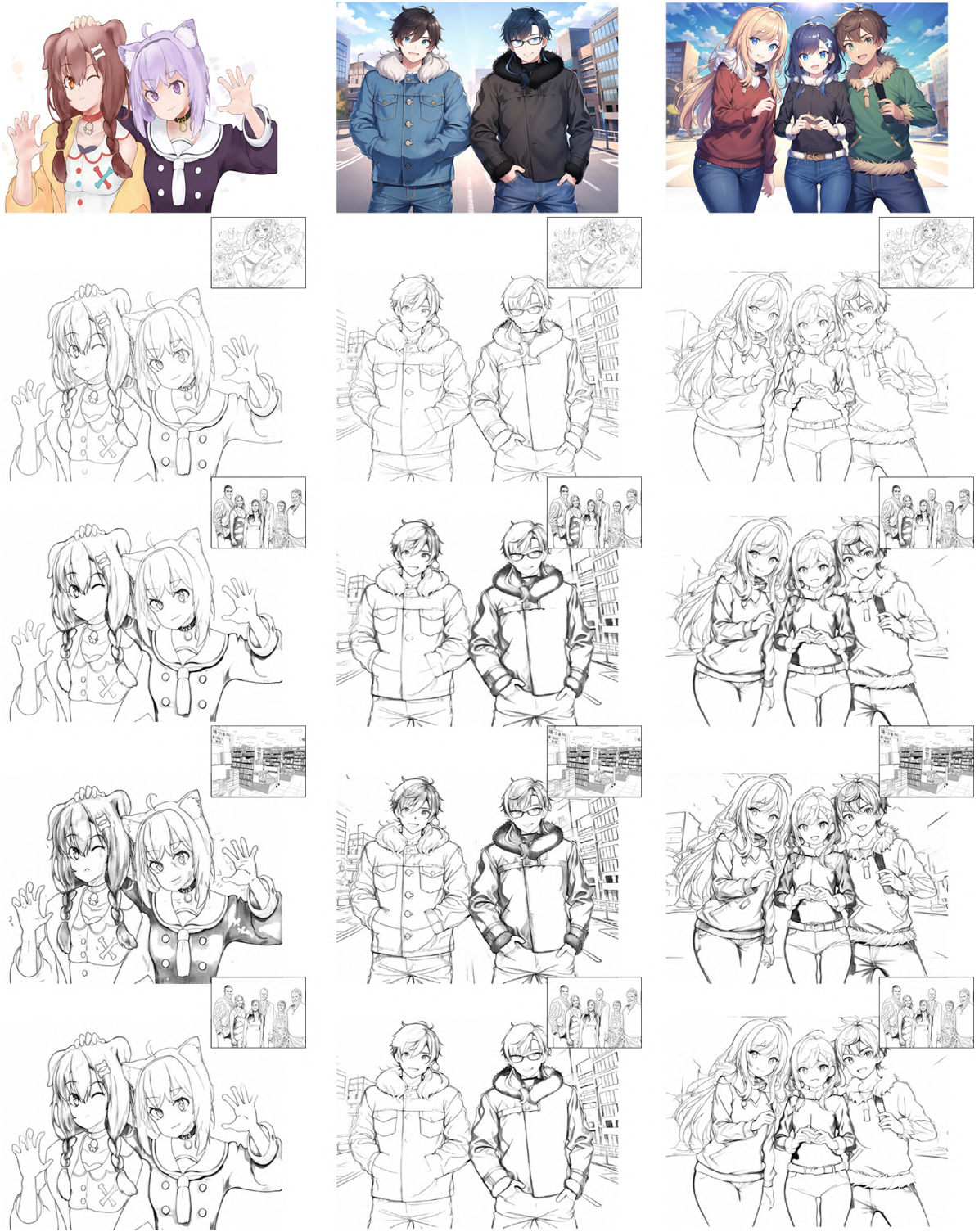

Figure 21: Additional examples of sketch extraction produced using our method. © Comete_atr